\documentclass[letterpaper, 10 pt, conference]{ieeeconf} 

\title{robot_formation_gen}
\author{Paper ID: }
\date{December 2019}
\IEEEoverridecommandlockouts                              

\usepackage[utf8]{inputenc}
\usepackage{bm}
\usepackage{amsmath}
\usepackage[Symbol]{upgreek}
\usepackage{graphicx}
\usepackage{graphics}
\usepackage[tight]{subfigure}
\usepackage{floatrow} 
\usepackage{multirow}
\usepackage{array}
\usepackage{enumerate}
\usepackage{sidecap}
\usepackage[all]{xy}
\usepackage{authblk}
\usepackage{comment}
\usepackage{changepage}   
\usepackage{wrapfig}
\usepackage{url}
\usepackage{tabularx,pbox,booktabs,caption}
\usepackage{flushend}
\usepackage{anyfontsize}
\usepackage{float}
\usepackage{todonotes}
\usepackage[percent]{overpic}
\graphicspath{{./figs/}}
\usepackage{amssymb}
\usepackage{amsfonts}
 \providecommand{\norm}[1]{\lVert#1\rVert}
\usepackage{color}
\usepackage{blindtext}
\usepackage[T1]{fontenc}
\usepackage[english]{babel}

\usepackage{amsthm}

\title{\LARGE \bf
Swarming of Aerial Robots with Markov Random Field Optimization
}

\author{Malintha Fernando, Lantao Liu
\thanks{M. Fernando and L. Liu are with Luddy School of Informatics, Computing, and Engineering  at Indiana University, Bloomington, IN 47408, USA. E-mail:
        {\tt\small \{ccfernan, lantao\}@iu.edu}.
}
}

\theoremstyle{remark}

\theoremstyle{definition}
\newtheorem{definition}{Definition}

\begin{document}

\maketitle
\section{Abstract}
\textbf{
Swarms are highly robust systems that offer unique benefits compared to their alternatives. In this work, we propose a bio-inspired and artificial potential field-driven robot swarm control method, where the swarm formation dynamics are modeled on the basis of Markov Random Field (MRF) optimization. We integrate the internal agent-wise local interactions and external environmental influences into the MRF. The optimized formation configurations at different stages of the trajectory can be viewed as formation ``shapes" which further allows us to integrate {dynamics-constrained motion control} of the robots. We show that this approach can be used to generate dynamically feasible trajectories to navigate teams of aerial robots in complex environments. 
}

\section{Introduction}

Formation controlling of multiple aerial vehicles in obstacle environments has been of significant importance and necessity in many application scenarios such as search and rescue, disaster response, infrastructure maintenance~\cite{cardona2019robot}.
Multiple attempts have been conducted in controlling large groups of such robots in complex environments \cite{honig2018trajectory} \cite{cappo2018online}. 
Even though these approaches show promise in accomplishing their defined tasks, a majority of multi-robot formation control applications move the robots through static patterns with fixed individual goal positions. Approaches based on the concurrent assignment and trajectory planning \cite{turpin2013trajectory} have been proposed to navigate robot formations through changing environments such as narrow corridors, still they transit the robots through static formation patterns defined over a static graph. It is apparent that such formations need to be resilient to the changes in the environment and thus comparable to biological swarms. However, most of the existing formation control mechanisms compute trajectories from predefined start to end positions. It is more desirable to treat the entire team of robots as a single cohesive entity and hence not to specify individual goal positions, but a single target for the team. In \cite{michael2011control} and \cite{desai2001modeling}, the authors define the robot swarm using a single group element, yet the approaches suffer from determining suitable robot formations for different segments of the swarm trajectory as the robots move. 



In this work, we model the robot swarm as a cohesive entity with pairwise interactions between neighboring robots. This allows us to navigate robots by mimicking biological swarm behaviors. In \cite{shishika2017mosquito}, the interactions between the robots have been modeled as a damped oscillator to mimic the attraction and repulsion forces between them. Yet, the behaviors are limited to the pursuit and obstacle-free environments. Various approaches based on artificial potential fields (APF) have been proposed to navigate robots in obstacle fields while generating natural like formations controlling \cite{ze2012formation} \cite{gayle2009multi}. However, their use has been limited for robots with first or second-order robot dynamics. We seek a set of non-collision paths for a group of robots in an APF with more general control and dynamics. More specifically, we propose to combine the capability of APF to generate cohesive swarm behaviors with trajectory optimization to compute safe and smooth trajectories for generic robots with complex dynamics. 

\begin{figure}
\centering
        \includegraphics[trim={0cm 5cm 0cm 4cm}, clip, width=\textwidth]{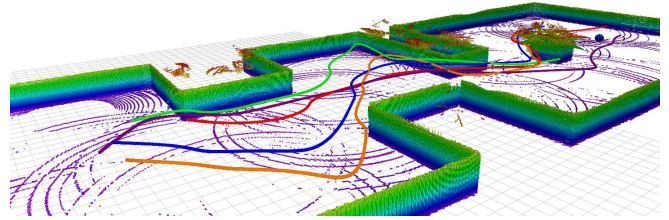}
        \caption{Trajectories of a team of five quadrotors moving through an obstacle field.}
        \label{fig:starlings}
\end{figure}

A common modeling approach of APF for robot navigation is to drive the robots toward a goal position, which is in general defined by a global minimum of the potential field. Further, MRFs provide a general energy minimization framework for pairwise and higher-order interactions, which are abundant in APF robot navigation paradigm. This inspired us to study the applicability of MRFs in robot swarm navigation. In contrast to other APF based approaches where the robots are directly subjected to a force vector field \cite{okubo1986dynamical}, \cite{ze2012formation}, we propose a novel approach to integrate potential fields into dynamic MRF optimization. First, we compute a set of synthesized APFs depending on the nature of the exertion of the forces (i.e., internal or external) on the swarm. Secondly, we incorporate the synthesized potential fields that contribute to the \textit{swarm energy} in the MRF energy function. Thirdly, the MRF is iteratively optimized by energy minimization constrained on the navigable spaces of the robots until the swarm energy is converged and subsequently the swarm has reached the goal. In other words, these converged swarm positions also represent the maximum a posteriori (MAP) distribution of the MRF. The results of each optimization iteration are used to construct a collision-free path for the robot swarm over an obstacle field. Finally, we smooth the resulting paths via trajectory optimization to generate navigable trajectories adhering to the non-holonomic behavior of the robots. Opposed to the existing APF approaches, our method allows the robot paths to be computed and evaluated before the execution,  so that further safety and collision checks can be performed.


\section{Related Work}

A vast amount of studies has been conducted in the field of path planning for multiple robots. We can distinguish the methods as graph/topology-based \cite{sanchez2002delaying} \cite{guo2002distributed} \cite{desai2001modeling} \cite{yu2016optimal}, artificial potential based \cite{chuang2007multi} \cite{nazarahari2019multi} \cite{bennet2010distributed} \cite{toksoz2019decentralized} and constrained optimization based \cite{augugliaro2012generation} \cite{alonso2017multi} approaches. In general, the graph-based approaches, e.g., probabilistic road maps (PRM), create a graph of the free space and then attempt to find collision-free paths for multiple robots. In \cite{augugliaro2012generation} and \cite{alonso2017multi} authors propose to use sequential convex programming (SCP) to evaluate the non-convex collision avoidance constraints of multiple robots during the trajectory optimization phase. A majority of approaches based on graph and constrained optimization, do not consider the group of robots as a single cohesive entity, hence each robot is required to have an individual start and goal positions. Even though \cite{alonso2017multi} showed that it is possible to perform swarm split and merge with SCP, the interim formations are required to be predefined. 
APF based methods show promising performances in terms of bio-mimicry in robotic systems with swarm splitting and merging during the navigation \cite{bennet2010distributed}. However, these methods are mostly limited the robot controllers to be in higher-order dynamics which must use higher-order dynamics.
\cite{toksoz2019decentralized} \cite{shishika2017mosquito}. Also, they require the robots to navigate in real-time to activate the potentials, prohibiting evaluating the planned trajectories prior to execution. Further, many APF based methods require to introduce external dissipative forces to the system to reach an equilibrium \cite{shishika2017mosquito}. In \cite{baras2004control} and \cite{xi2006gibbs}, the authors proposed a similar framework, based on MRF with Gibbs sampling, yet the connection to underlying dynamics of the robots have not been considered. Since our method is compatible with the \textit{receding horizon planning} (RHP) scheme, which lies in between the real-time planning execution and offline planning, the robot trajectories can be further optimized and evaluated prior to the execution. In \cite{spears2004distributed} the authors show that collective behavior can result in patterns such as pentagons and squares. We show that similar results can be achieved via our method in the absence of other potential fields. 


\section{Preliminaries}

Let $\mathcal{G} = (\mathcal{V}, \mathcal{E})$ be an undirected graph with vertices $\mathcal{V} = \{1,2,...,N\}$ and edges $\mathcal{E} \subset \{\mathcal{V} \times \mathcal{V}\}$. Each vertex, $i \in \mathcal{V}$ is associated with a latent random variable $\textbf{x}_i$ where $\textbf{x}_i$ is chosen from a possible labelling set $X_i$. In this work, we use navigable positions of robots to define the labelling set. In MRF, absence of edges between any two given nodes denotes the conditional independence between them. We define a \textit{Markovian blanket} $\mathcal{N}_i$ of a node based on the local conditional independence of $\mathcal{G}$ as, 

\begin{equation}
\label{local_m}
    P(\textbf{x}_i | \{\textbf{x}_j\}_{j \in \mathcal{V} \backslash i}) = P(\textbf{x}_i |  \{\textbf{x}_j\}_{j \in \mathcal{N}_i}).
\end{equation}

In the upcoming section, we compute a graph depending on the Markovian blanket of a given robot. We base our MRF on this graph and perform energy minimization on its \textit{maximal cliques} factorization. 

\begin{definition}
\label{clique_def}
A \textit{clique} $c$ of an undirected graph $\mathcal{G} = (\mathcal{V}, \mathcal{E})$ is a subset of vertices, $c \subseteq \mathcal{V}$ where all the vertices are adjacent to each other. A \textit{maximal clique} is a clique which cannot be extended by including one more adjacent vertex, that is, a clique which does not exist exclusively within the vertex set of a larger clique.
\end{definition} 

With this definition, we introduce a clique factorization for the MRF. The joint probability distribution of any MRF can be decomposed into a product of \textit{clique potential functions} $\psi_c( \mathrm{\textbf{x}}_c)$ as follows.

\begin{equation}
\label{h-c}
    p(x) = \frac{1}{Z}\prod_{c \in C}(\psi_c( \mathrm{\textbf{x}}_c)).
\end{equation}

Here, $c$ is a maximal clique, $c \in C$ and $\mathrm{\textbf{x}}_c$ is a labelling of clique $c$, $C$ is the set of all maximal cliques and $Z$ is a \textit{partition function}.

\subsection{Energy Minimization on Markov Random Fields}
By restricting the value of $\psi_c( \mathrm{\textbf{x}}_c)$ to be strictly positive for any \textbf{x}$_c$, we can express potential functions as exponentials,
\begin{equation}
\label{Ex}
    \psi_c (\mathrm{\textbf{x}}_c) = \exp{(-E(\mathrm{\textbf{x}}_c))},
\end{equation}
\begin{equation}
\label{p in E}
    p(x) = \frac{1}{Z} \exp{(\sum_{c} -E(\mathrm{\textbf{x}}_c))},
\end{equation}
where, $E(\mathrm{\textbf{x}}_c)$ is called an \textit{energy} function. We compute the energy of a clique given the labelling using a set of superimposed artificial potential functions. The partition function $Z$ can be defined as follows to ensure the distribution sum to 1~\cite{bishop2006pattern}.
\begin{equation}
\label{Z}
    Z = \sum_{\mathrm{\mathbf{x}}} \prod_{c \in C}\exp({\psi_c(\mathrm{\textbf{x}}_c)})
\end{equation}

In the following section, we discuss our approach to generate strictly positive clique energies, $E(\textbf{x}_c)$. Therefore, by minimizing the energy function, we can obtain Maximum a Posteriori (MAP) label distribution $\mathrm{\textbf{x}}^*$ for the MRF according to \eqref{Ex} and \eqref{p in E}. We use straightforward Iterated Conditional Modes (ICM)~\cite{bishop2006pattern} for the optimization phase of the MRF in this work. 

\section{Approach}

\subsection{Bio-Inspired Interaction Graph}

\begin{figure}
\subfigure[]
 	{\label{fig:interaction_5} \includegraphics[scale=0.44,trim={4cm 2.3cm 3.8cm 2cm}, clip]{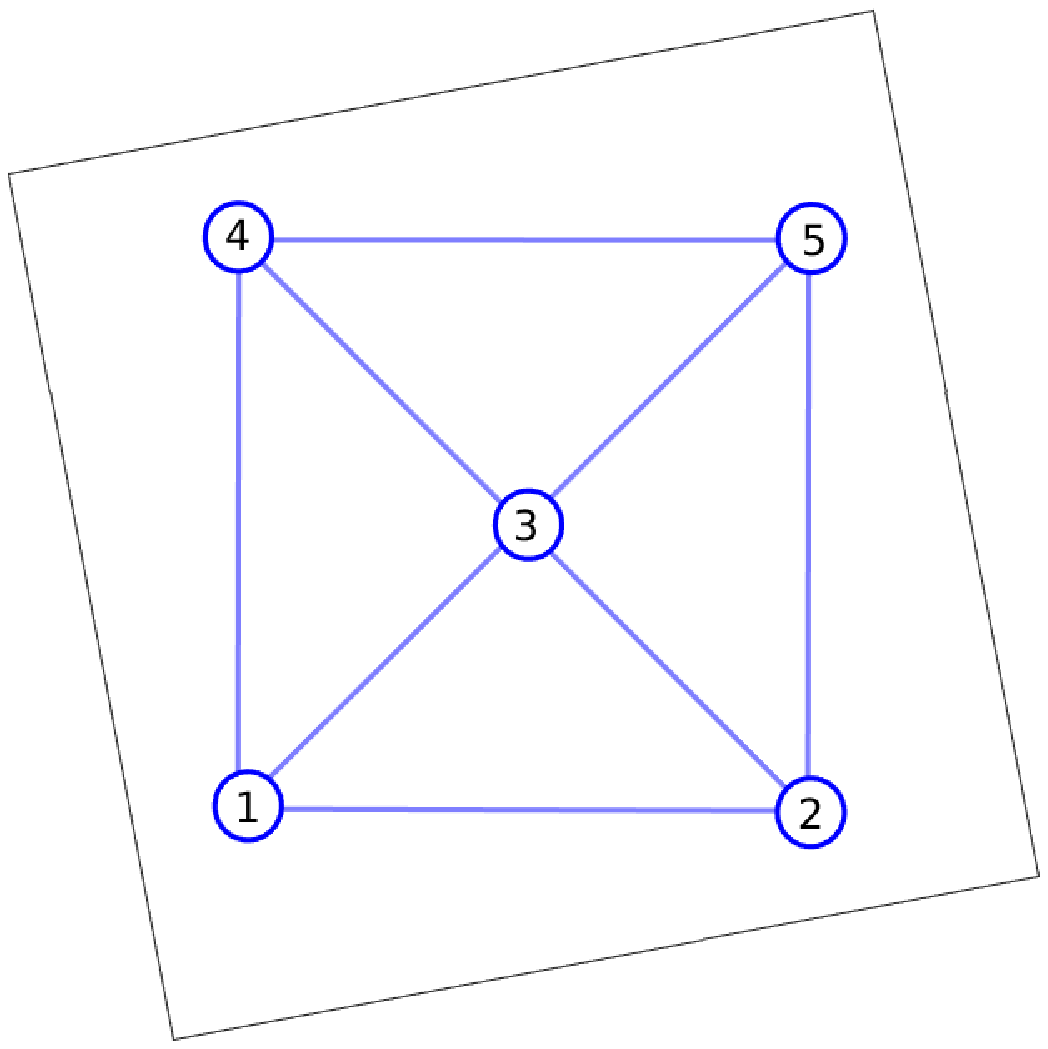}}
\subfigure[]
 	{\label{fig:factors} \includegraphics[scale=0.31,trim={0cm 0cm 0cm 0cm}, clip]{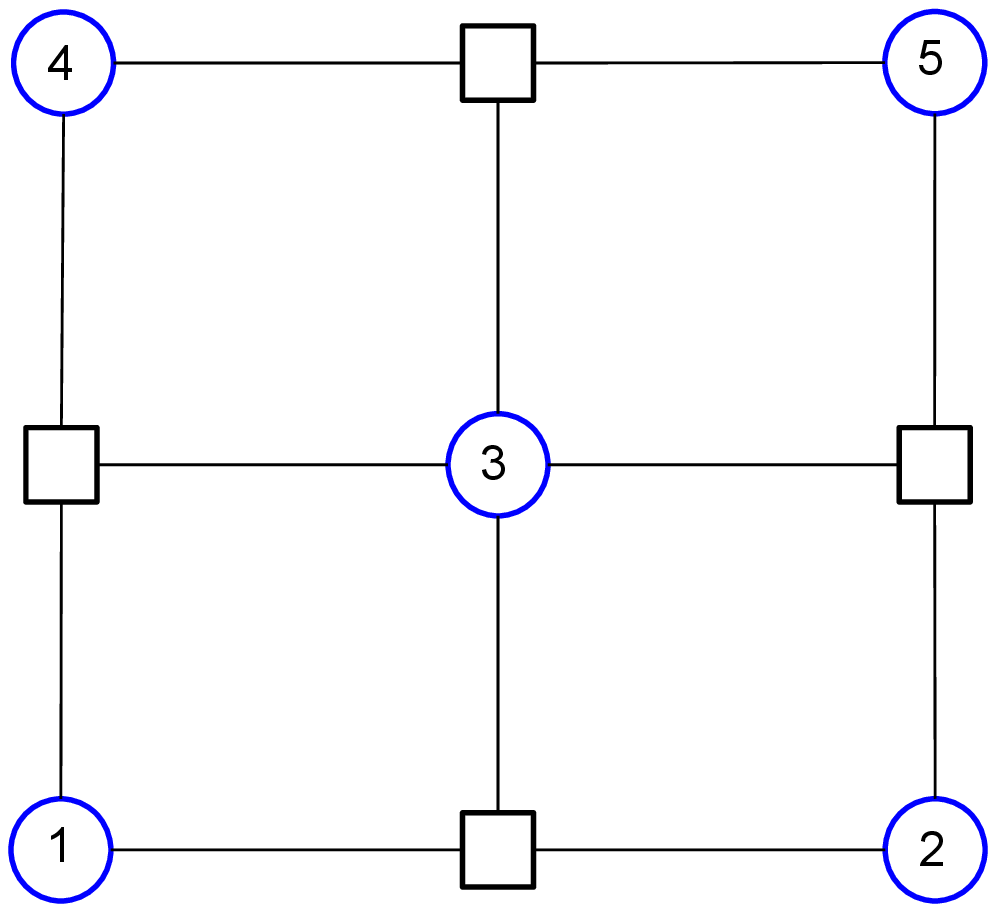}}
\caption{(a) The interactions graph for $k=3$ scenario. (b) Factor graph representation of the maximal cliques of $\mathcal{G}$.}
\end{figure}

\begin{figure*}[ht]
  \centering
  \subfigure[]
  	{\label{fig:5_robs} \includegraphics[scale=0.3,trim={1cm 1cm 1cm 0.5cm}, clip]{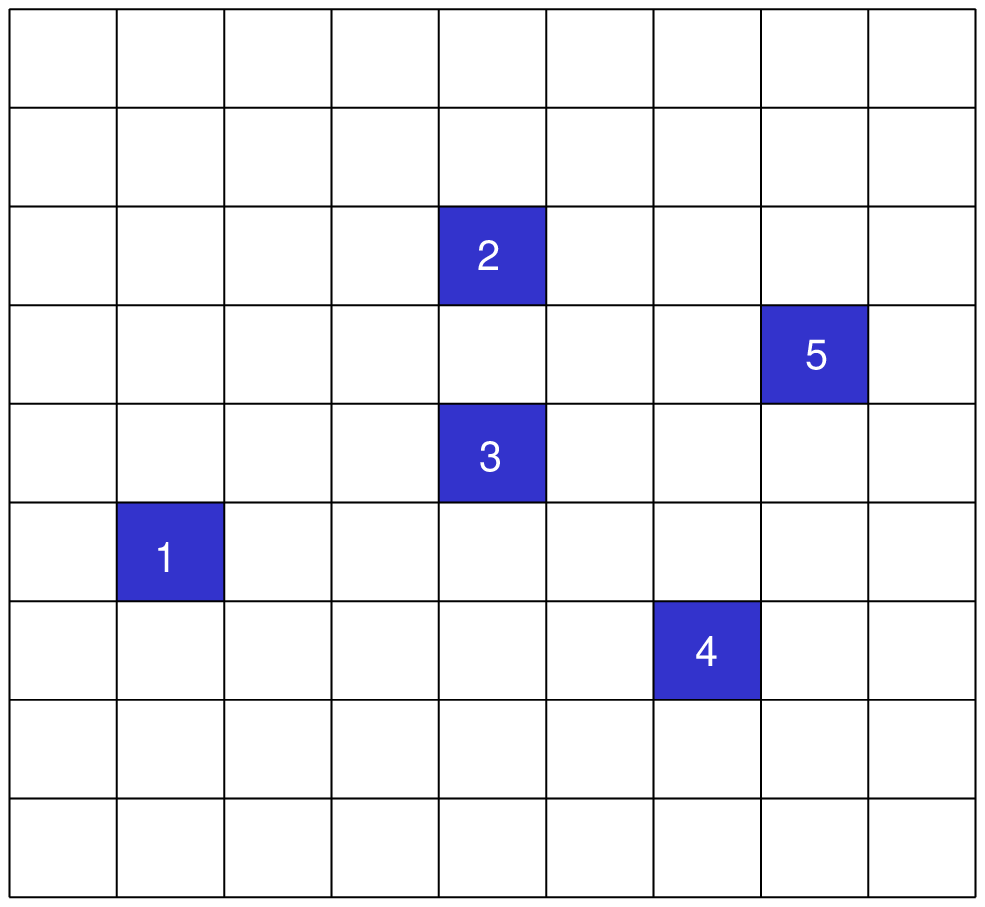}}
  \hfill
  \subfigure[]
 	{\label{EI} \includegraphics[scale=0.38,trim={2cm 2cm 2cm 3cm}, clip]{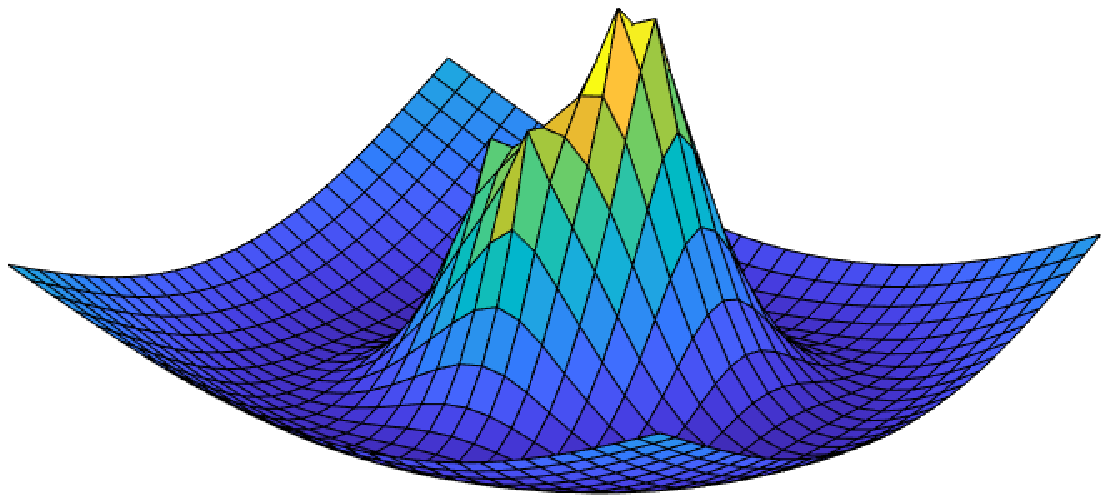}}
    \hfill
\subfigure[]
 	{\includegraphics[scale=0.32,trim={2.5cm 2cm 2.5cm 2.5cm}, clip]{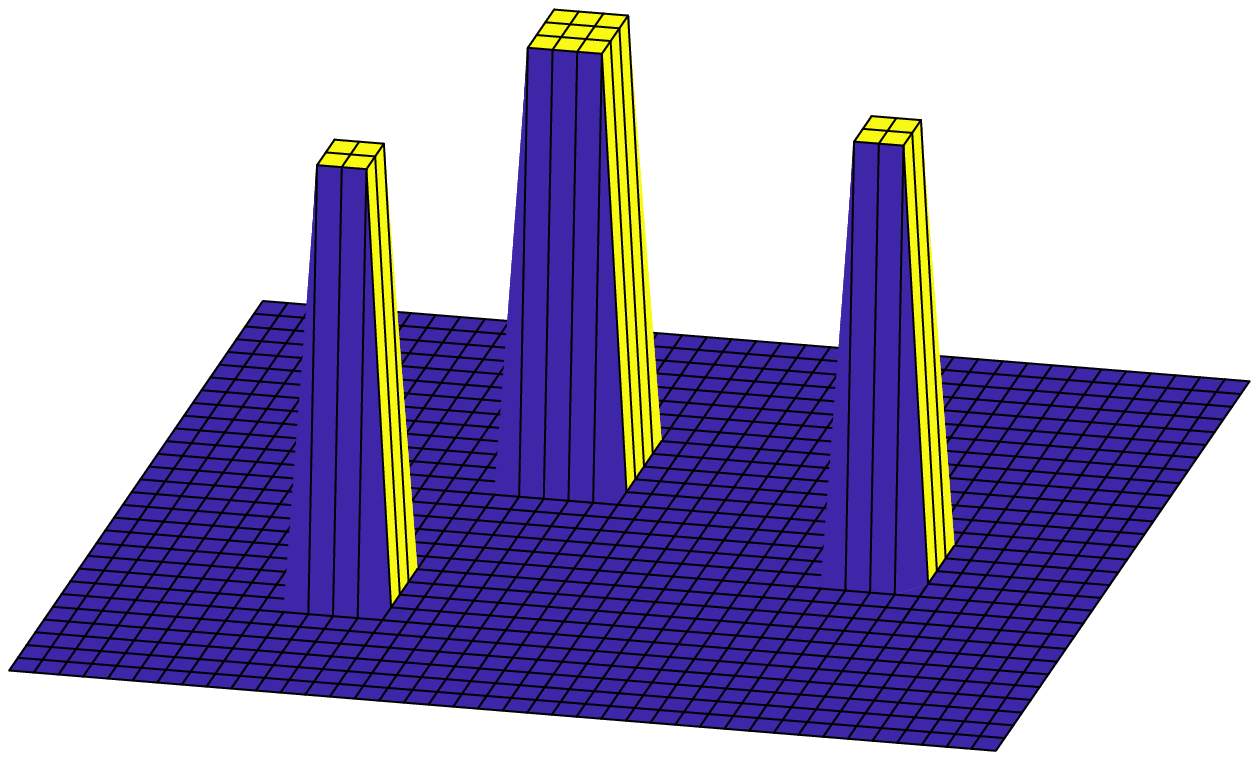}}
 	\hfill
\subfigure[]
 	{\includegraphics[scale=0.32,trim={2.5cm 2cm 2.5cm 2cm}, clip]{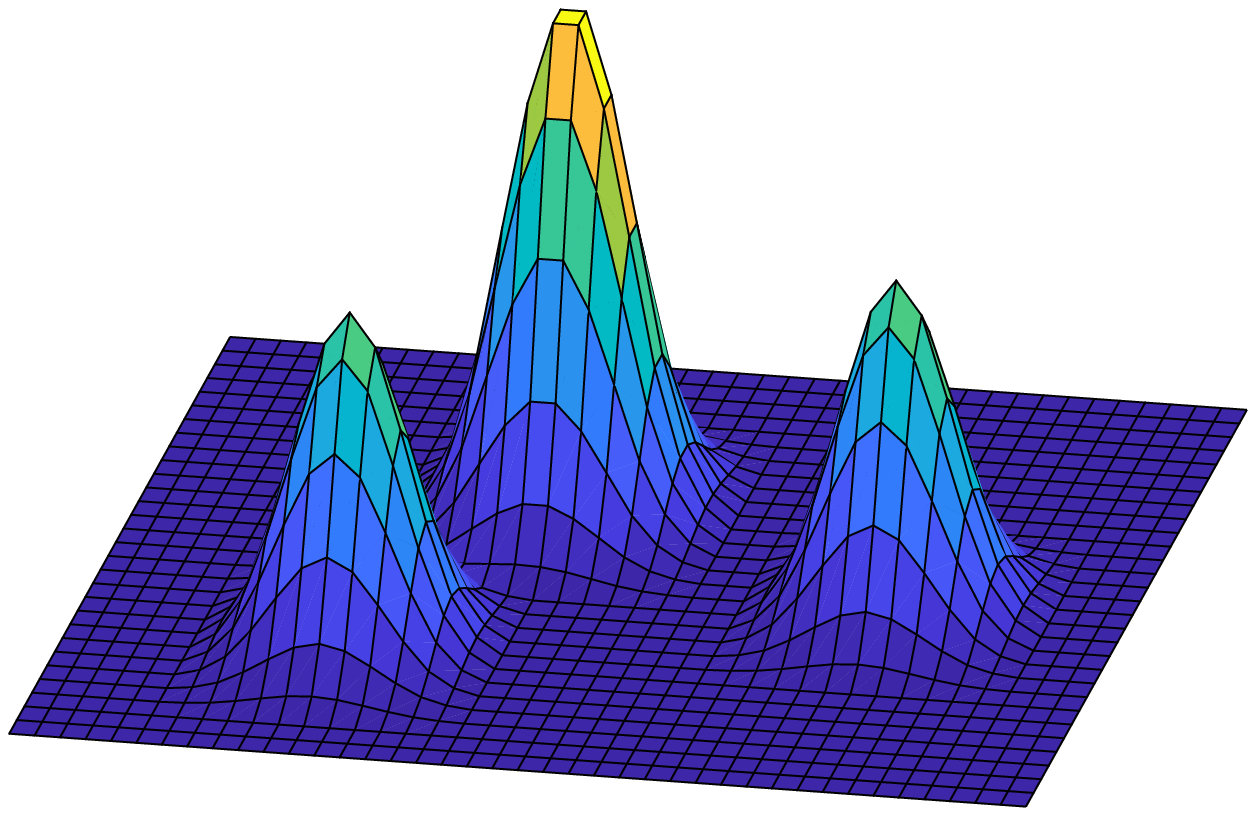}}
\caption{(a). A distribution of 5 interacting robots. (b) The corresponding interaction potential map. (c) The obstacle map ($\hat{m}$). (d) Smoothed obstacle potential field.}
\label{goal_pot}
\end{figure*}

Studies of the emergence of large flocks of animals show that collective behavior follows a set of simple rules which define the behavior of individual members. A field study of \textit{murmurations} of European Starlings, \cite{ballerini2008interaction} suggests that the social ruling of such behaviors depend on the topological neighborhood rather than the metric distance. We use this assumption to model natural like interactions between the robots. In this work, we consider the neighborhood of a robot consists of the nearest $k$ robots, $k \leq N$, where $N$ is the number of robots in the swarm. The occupancy of a given location by a robot in the environment can be represented by an occupancy grid map, and as a first step we assume there is no uncertainty for the localization of robots or obstacles.
Fig. \ref{fig:5_robs} shows a distribution of 5 robots on a two dimensional environment discretized into a $10 \times 10$ grid map. As we consider an undirected graphical model to represent the interactions, a bidirectional edge is constructed for any two interacting robots. An interaction graph $\mathcal{G} = (\mathcal{V}, \mathcal{E})$ constructed for the given distribution is illustrated in Fig. ~\ref{fig:interaction_5}. For the clarity of representation, we illustrate a $k=3$ scenario here. The set of vertices $\mathcal{V}$ and the set of edges $\mathcal{E}$ represent the interacting robots and their pairwise connections, respectively. The corresponding factor graph for $\mathcal{G}$ is depicted in Fig. ~\ref{fig:factors}, where $\{c_i\} \in C$ is a maximal clique in the factorization. 

\begin{definition}
\label{swarm_energy}
Given labelling \textbf{x} for all the vertices $\mathcal{V}$, clique labelling $\textbf{x}_c$ where $\textbf{x}_c \subset \textbf{x}$, we define \textit{swarm energy} $\varepsilon$ as the summation of all maximal clique energies of the interaction graph.
\begin{equation}
    \varepsilon(\textbf{x}) = \sum_{c} E_c(\textbf{x}_c)
    \label{swarm_en_sum}
\end{equation}
\end{definition} 
Following the above definition, total \textit{swarm energy} for the group of robots in Fig. ~\ref{fig:5_robs} can be written as $\varepsilon(\textbf{x}) = \sum_{i=1}^4 E_i(\textbf{x}_i)$. Due to the summation, we require all the clique energies to be negative to avoid canceling each other. According to \textit{local Markovian property} in \eqref{local_m}, we define Markovian blanket of robot $i$, $\mathcal{N}_i$ to be the same as the local neighborhood of a robot $i$. Further, it is clear from Fig.~\ref{fig:interaction_5}, that the level of the conditional independence of the graph depends on the local neighborhood of individual robots and their spatial distributions. As we compute the total swarm energy as a function of the clique factorization of $\mathcal{G}$, our approach is independent of the level of Markovianity of the \textit{interaction graph}. However, in order to navigate the robots as a cohesive entity, we impose a necessary and sufficient condition for $\mathcal{G}$ as, $\forall i \in \mathcal{V}, \hspace{0.25cm} \mathcal{N}_i \neq \varnothing$.


In other words, we restrict the initial interaction graph not to  consist of completely independent vertices. In practice, this can be achieved by placing the robots in such a way that the distances between them not to exceed the limits of their communication ranges. This will let the robots to establish neighborhoods at the time of initialization.

\subsection{Construction of Artificial Potential Fields}

Artificial potential fields have been widely used in robot navigation literature for maneuvering robots with varying underlying controllers. We extend the same concept to model static and dynamic obstacles in the environment to compute the paths for groups of robots of finite size. As complex environments consist of components with varying levels of interests for the robots, we construct multiple potential functions to represent distinctive interactions. In this section, we propose methods to compute potential functions for obstacles, neighboring robots and a goal position that the group of the robots need to reach. As the potential field of a potential function $f(q): \mathbb{R}^n \xrightarrow{} \mathbb{R}$, is simply the vector field over all $q \in \mathbb{R}^n$, we may use two terms interchangeably. 

In natural or artificial physical systems, \textit{potential energy} can result from various interactions that occur within and amongst the systems. 
In this work we presume that the robots are attracted to a predefined goal position $p_g \in \mathbb{R}^2$ and repelled by the obstacles in the environment. Therefore, we construct two potential fields, \textit{goal potential} and \textit{obstacle potential}, as external potentials that will be exerted on the swarm. 
Further, we construct an interaction potential field to model the internal interactions between the robots as they navigate.

\subsubsection{Interaction Potential}

We employ {\em Morse potential function}\cite{gazi2013lagrangian} to model the \textit{interaction potential}, $\rho_I$ between robots.
The function has been widely studied to model the interactions between self-propelled particles in the literature \cite{d2006self} \cite{gazi2013lagrangian}. The repulsion and the attraction components of the function control the output energy given the distance between the agents. As the distance between agents decreases to some characteristic length scale, the repulsion term dominates the attraction term which allows for the cohesion of the swarm on large scales while creating space between robots locally. However, a minimum energy setting can be found for any pairwise interaction, which in large groups of robots can be identified as the global minimum in the absence of other potentials.
For any two neighboring robots, $i$ and $j$ at $p_i$ and $p_j$, we define the interaction energy $\rho_I$ as follows. 
\begin{equation}
    \rho_I(p_i, p_j) = \rho'_I(p_i, p_j) + \zeta,
\end{equation}
\begin{equation}
    \label{morse}
    \rho'_I(p_i, p_j) = -a_I\exp{(-\norm{\frac{d_{ij}}{k_a}})} + b_I\exp{(-\norm{\frac{d_{ij}}{k_r}})},
\end{equation}
where, $a_I,b_I,k_a,k_r > 0$, $b_I \geq a_I$, $k_a > k_r$, $\frac{b_I k_r}{a_Ik_a} < 1$ and $d_{ij} = (p_i - p_j)$. The parameters $a_I$,$b_I$,$k_a$ and $k_r$ are bounded as such to define the local minimum behavior of the Morse potential function \cite{gazi2013lagrangian}. We introduce a constant $\zeta$ to restrict the interaction potential to be strictly positive. Therefore,
\begin{equation*}
    \zeta > \inf(\rho'_I(p_i, p_j)).
\end{equation*}
Let $d_{min}$ be the distance that gives the minimum interaction energy.
\begin{equation}
    \zeta > |\rho'_I(d_{min})|.
\end{equation}
By considering the partial derivative of \eqref{morse} w.r.t $\norm{d_{ij}}$,
\begin{equation*}
    d_{min} = \ln{(\frac{a_Ik_r}{b_I k_a})}(\frac{k_r k_a}{k_a - k_r}).
\end{equation*}
Fig.\ref{EI} shows an interaction potential field for a group of 5 robots. The valley around the given arbitrary robots formation proves that there is a globally minimum energy robot configuration when the robots are not subjected to any other potential energy.

\subsubsection{Goal Potential}
We define a static and strictly positive \textit{goal potential} function, $f_e: \mathbb{R}^2 \xrightarrow{} \mathbb{R}$ that attracts the robots toward the goal, $p_g$ over the environment. As we seek a minimum energy setting for the robots, we construct the potential energy with a global minimum at the goal region. For any $p \in \mathbb{R}^2$, the goal potential $\rho_G(p)$ can be defined as,
\begin{equation}
    \rho_{G}(p) = a_G\exp({\frac{\norm{p - p_g}}{k_G}}),
\end{equation} 
where $a_G, k_G > 0$. By defining the goal potential function as above, we can generate a highly attractive potential toward the goal when the robots are further away, resulting the robots to navigate toward the goal for minimizing the energy. 

\subsubsection{Obstacle potential}

As the robots are driven toward the goal, we use an \textit{obstacle potential} function to navigate the robots away from the obstacles. A multitude of methods has been proposed to compute potential fields for obstacles. A widely regarded approach for computing the obstacle potential fields suggested by \cite{khatib1986real}, using the shortest distance to the obstacle border. However, the method requires calculating the distances from a given point in the environment to the nearest border of the obstacle, making this approach can be cumbersome when calculating the potential fields for complex obstacle shapes. We propose using a high-pass filter (HP) and a Gaussian kernel function on the occupancy grid map as an alternative approach. By using an occupancy grid map $m$ of the environment, we construct a secondary map $\hat{m}$, by applying a high pass filter as in \eqref{m_hat}. The HP filter essentially separates the obstacle and free space by thresholding. We apply a Gaussian kernel on the secondary map resulting a smoothly elevated and strictly positive obstacle potential field $\rho_O$. For a given coordinate $p = (x,y)$ and $P(m_{xy})$ we define the secondary map,
\begin{equation}
    \label{m_hat}
    \hat{m}_{xy} = 
    \begin{cases} 
        \hspace{10pt} \Gamma & P(m_{xy}) \geq 0.5, \hspace{10pt} \text{Where } \Gamma > 0 \\
        \hspace{10pt} 0 & \text{Otherwise}.
    \end{cases}
\end{equation}

We apply a Gaussian kernel ${\textbf{G}}$ on $\hat{m}$ to obtain the obstacle potential field. The energy of $p$ on the potential field is defined as,
\begin{equation}
    \rho_{O}(p) = \hat{m}_{xy} \circledast {\textbf{G}},
  \label{obs_poten}
\end{equation}
where, $\circledast$ is the convolution operator.

\begin{figure}
 	{\includegraphics[scale=0.5,trim={1cm 0cm 0cm 0cm}, clip]{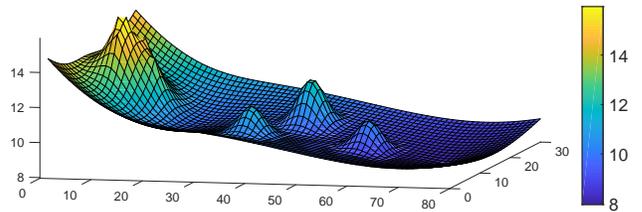}}
\caption{The synthesized environment potential field .}
\label{all_pot}
\end{figure}

By synthesizing the goal, obstacle and interaction potential fields, we generate a \textit{combined potential field}, $\rho_{com}$ such that $\rho_{com} = \rho_I + \rho_O + \rho_G$. As a result of combining with the interaction potential, the synthesized potential field dynamically changes as the robots move in the environment. 
This property can be used to search for collision free paths for robots over the environment using an iterative and local search. Fig.~\ref{all_pot}, shows a combined potential field. The valley to the right of the map represents the goal position at $(65,15)$. The higher energy area represents the starting position of the robots and their interaction energies in a global perspective.

\subsection{Markov Random Field Optimization}

 \begin{figure*}[ht] \vspace{-10pt}
  \centering
  \subfigure[]
 	{\includegraphics[width=0.18\textwidth,trim={1cm 1cm 1cm 0.5cm}, clip]{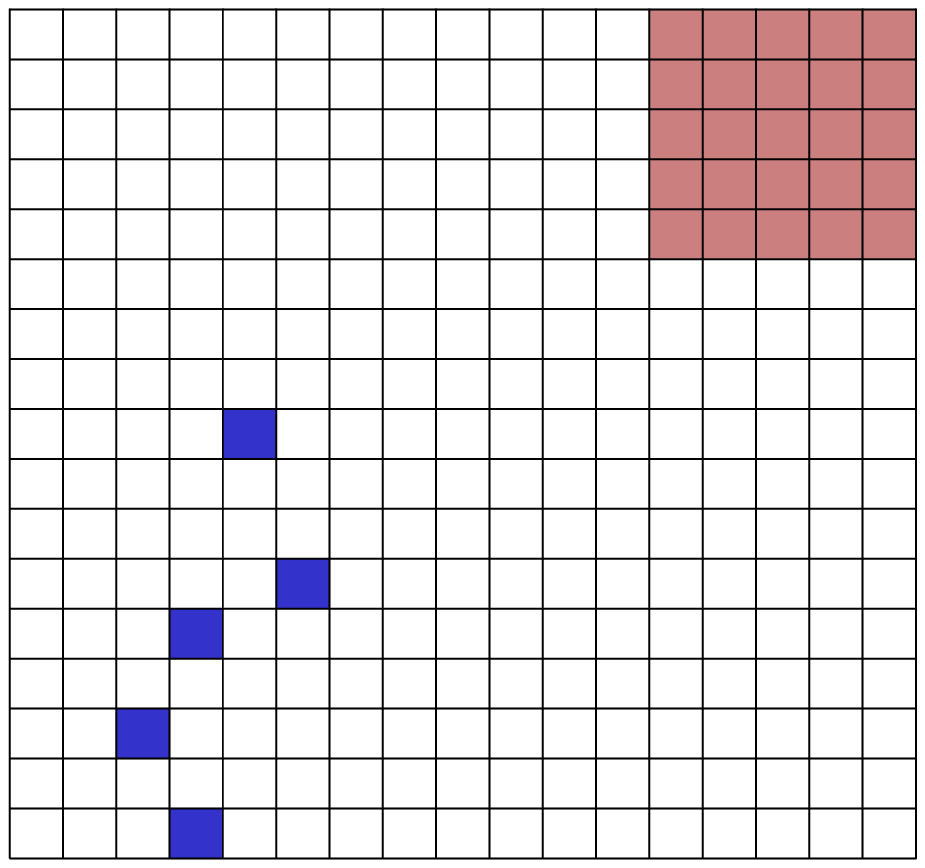}}
    \hfill
\subfigure[]
 	{\includegraphics[width=0.18\textwidth,trim={1cm 1cm 1cm 0.5cm}, clip]{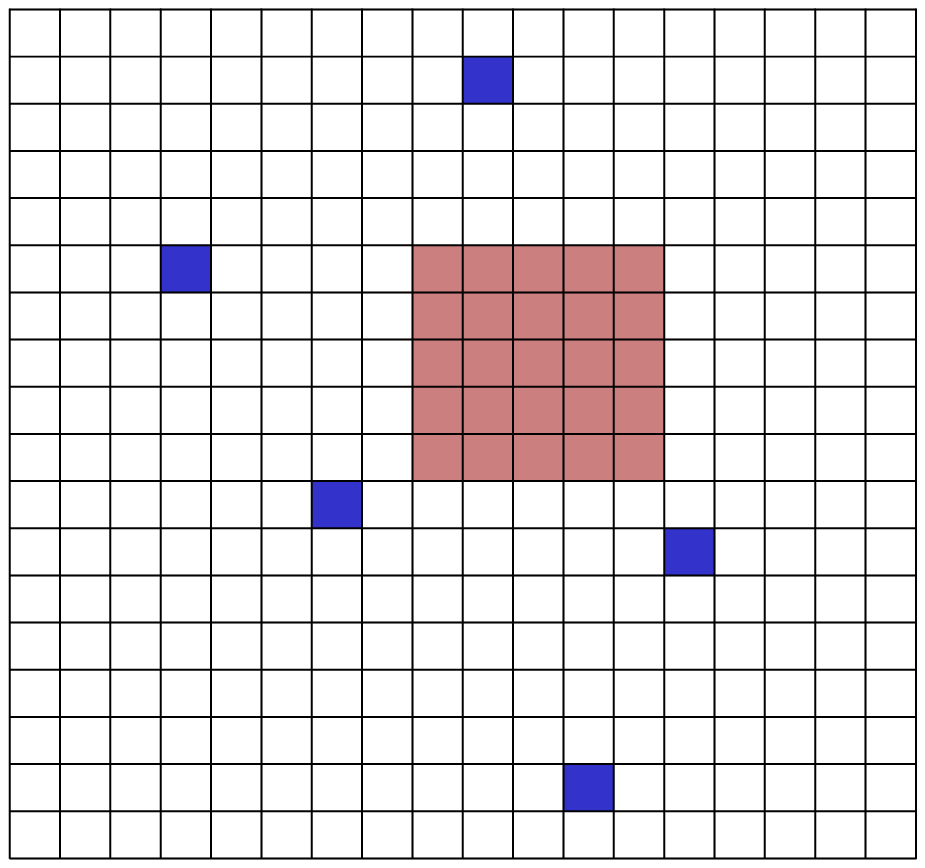}}
 	\hfill
\subfigure[]
 	{\includegraphics[width=0.18\textwidth,trim={1cm 1cm 1cm 0.5cm}, clip]{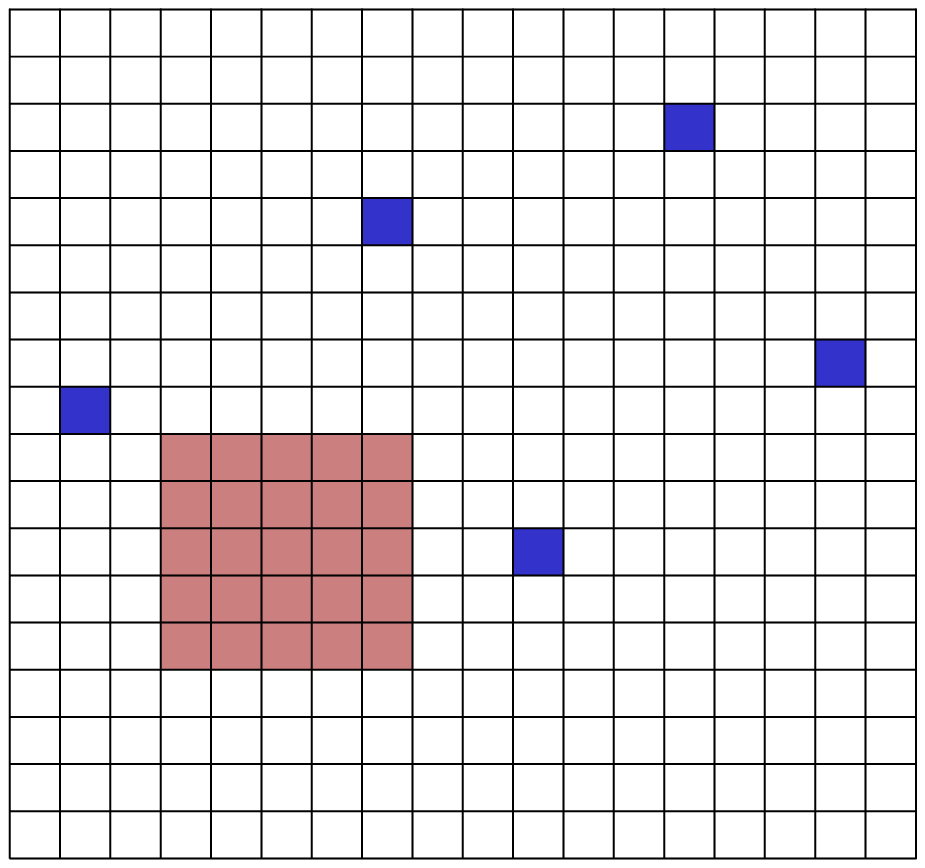}}
 \hfill
 \subfigure[]
 {\includegraphics[width=0.18\textwidth, trim={1cm 1cm 1cm 0.5cm}, clip]{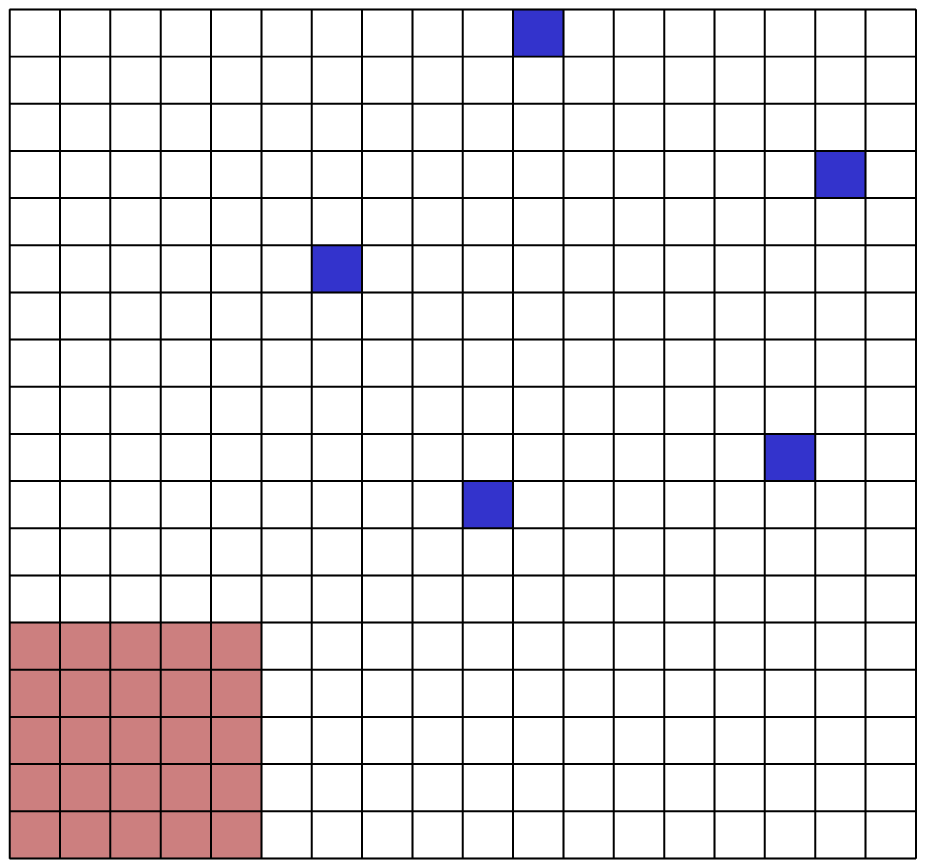}}
\hfill
 \subfigure[]
 {\includegraphics[width=0.20\textwidth, trim={2cm 1cm 1cm 0.5cm}, clip]{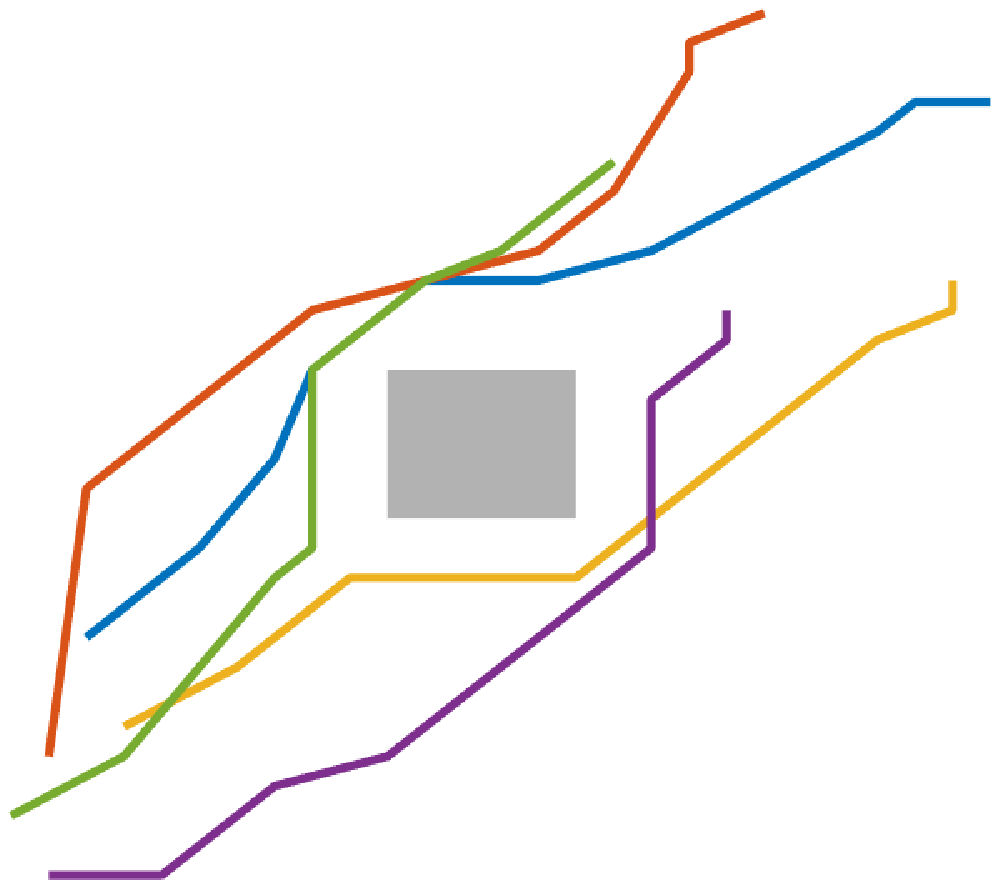}}
 
\caption{MRF optimization steps for 5 robots with a single obstacle. (a) Initial robot distribution. (b) (c) (d) After 4, 7 and 14 optimization steps. (e) Resulting discrete trajectories after waypoint pruning.}\vspace{-10pt}
\label{mrf_stages}
\end{figure*}


We construct a dynamic Markov random field using the interaction graph $\mathcal{G} = \{\mathcal{V,E}\}$ and $|\mathcal{V}| = N$, where each vertex $i \in \mathcal{V}$ is associated with a latent random variable $\textbf{x}_i^t \in X_i^t$. We define $X_i^t$ be a \textit{local search space}, which we use to obtain the latent values or labeling of the nodes iteratively.  Here, we use $t$ to denote each local optimization step. As robots navigate over the environment, we update the interaction graph and thus the random field. The local search space for a given robot, $X_i^t$ is also updated iteratively, hence each optimization step $t$ results in $N$ locally optimum position nodes that eventually construct a set of collision free discrete paths. In other words, we search for the global minimum of the MRF energy via local search. As $t$ grows, we obtain the latent values that minimizes the \textit{clique potentials}, Eq. (\ref{Ex}) and subsequently MAP distribution, $\textbf{x}^*$ for the MRF. Assuming a robot $i$ at $p_i^t \in \mathbb{R}^2$ at $t$-th iteration, we define $X_i^t$ as the general \textit{n-th order homogeneous neighborhood} of $p_i$ in the underlying occupancy grid map. Let $p_i = (x_i, y_i)$. Therefore, for any cell $p_j = (x_j, y_j) \in m$,
\begin{equation*}
    X_i^t = \{(x_j, y_j) \in m : (x_i - x_j)^2 + (y_i - y_j)^2 \leq n\}.
\end{equation*}
For a group of $N$ robots, the \textit{combined local search space} at time $t$ can be defined as $ X^t = \bigcup_{i=1}^{N}X_i^t.$ This \textit{combined local search space} can be used to perform MRF optimization via Iterated Conditional Modes (ICM) to generate collision free paths for multi robot teams. The order of the MRF increases with $k$ and the computational complexity grows with the order of the local search space, $n$. As we formulate this as optimization of a higher-order MRF with multiple labeling, \textit{graph cut} based optimization techniques would only be fit into a subset of the possible MRF models \cite{fix2011graph} \cite{kohli2008partial} \cite{ishikawa2010transformation}. As ICM is a general approach for optimizing MRF regardless of the order, we adopt it to compute the solution. ICM iteratively searches for the locally optimum label $\textbf{x}^{*t}_i$ for node $i$ that minimizes the clique energy $E_c$ by letting the other nodes to be stationary. Therefore, the size of the search space $X^t$ heavily affects the computational complexity. As $X^t$ directly depends on the order of the search space, the time complexity increases with $n$. We can limit the size of the search space by introducing heuristics at different stages of the trajectory. These heuristics can be used to avoid any further collisions and to control the directions that the robots are permitted to navigate. Let us consider the following properties to hold for inter-robot collision avoidance;
\begin{itemize}
    \item No two robots share the same position in the occupancy grid at the same time-step, $p_i \neq p_j, \forall i,  \forall j \neq i$.
    \item Robot paths may not intersect in between two adjacent time-steps.
\end{itemize}
We integrate the above properties into heuristics as, (a) checking the current robots positions in the local search space and removing any occupied cells and, (b) removing any duplicate entrees in the combined search space such that, for any two robots $i,j$, $X^t_i \cap X^t_j = \varnothing$. The search spaces can be further modified to refrain the robots from navigating away from the goal by trimming the preceding half.

We define two superimposed potential fields, $\rho_S$ and $\rho_D$, namely static and dynamic, by combining the previously defined potential functions. In this work, we let, $\rho_S = \rho_G + \rho_O$ and $\rho_D = \rho_I$. Following \eqref{Ex}, clique energy function is defined $E_c (\textbf{x}_c)$ for any $c \in C$ for $t$-th iteration,

\begin{equation}
\label{energy}
    E_c(\textbf{x}^t_c) = \sum_{i \in c}\rho_S(\textbf{x}_i^t) + \sum_{ij \in c}\rho_D(\textbf{x}_i^t, \textbf{x}_j^t),
\end{equation}
where $\textbf{x}_i^t \in X_i^t$, $\textbf{x}^t_c = \{\textbf{x}_i^t: 1<i<m\}$ and $m=|c|$.

By performing optimization of the MRF over combined local search space $X^t$ iteratively until the swarm energy $\varepsilon$, ~\eqref{swarm_en_sum} is converged, we obtain a series of nodes that can be used as waypoints for individual robots. A resulting path for robot $i$ can be denoted as $(\textbf{x}_i^0, \textbf{x}_i^{1*}, \dots , \textbf{x}_i^*)$, where $\textbf{x}_i^0$ is the starting position for the robot, $\textbf{x}_i^{t*}$ is the locally optimum labelling for the timestep $t$ and $\textbf{x}_i^* \in \textbf{x}^*$.

A waypoint pruning method can be applied to the resulting discrete trajectories after a fixed number of local MRF optimization steps (\textit{planning horizon}) to smooth the paths prior to the trajectory optimization. A simple waypoint pruning method would be to connect the endpoints of each resulting trajectory, unless the pruned path intersects with an obstacle or another robot's trajectory. This could be performed iteratively to identify the waypoints that can be pruned over the discrete trajectories. 

\subsection{Smooth Trajectory Generation for Receding Horizon Planning}

We adopt minimum snap polynomial trajectory generation introduced in \cite{richter2016polynomial} to optimize the resulting discrete trajectories. The trajectory optimization is performed iteratively on computed pruned paths till the robots converge to the goal position. The time allocation to navigate between the waypoints plays a crucial role here, as too large times can lead the resulting smooth trajectories to overshoot and highly deviate from the computed discrete paths. Therefore, we start with an initial guess for the time allocations, depending on a constant velocity traversal time. Further, as we considered waypoints, $\textbf{x}_i^t \in \mathbb{R}^2$ during the planning phase, we assume that the altitude of the drones are fixed to a constant throughout the trajectories. Let $H \leq t^*$ be the number of path nodes optimized at once where $t^*$ is the number of iterations that MRF optimization stage takes to fully converge.

Let $\omega \in \mathbb{R}^2$ be an arbitrary waypoint, and $ \textbf{t}_\textbf{k}$ be the corresponding timestep, where $\textbf{k}=(0,1..,H)$. The optimal polynomial trajectory of degree $d$ that goes through $H$ nodes can be defined as, $\sigma(t) = [\sigma_X(t),\sigma_Y(t)]$. Without loss of generality,
we use $\sigma(t)$ to denote an optimal trajectory. The $q$-th time derivative of $\sigma(t)$ can be written as the following piecewise,
\begin{equation}
\label{piecewise}
\frac{d^{q}}{dt^{q}}\sigma(t) = 
\begin{cases}
    \hspace{10pt} \sum_{j=0}^{d-q}{\alpha_{j1}t^j}  & \text{$t_0 \leq t < t_1$},\\
    \hspace{10pt} \sum_{j=0}^{d-q}{\alpha_{j2}t^j}  & \text{$t_1 \leq t < t_2$},\\
    \hspace{35pt} \vdots & \\
    \hspace{10pt} \sum_{j=0}^{d-q}{\alpha_{jH}t^j}  & \text{$t_{H-1} \leq t < t_H$},\\
\end{cases}
\end{equation}
where $\alpha_{j\textbf{k}} \in \mathbb{R}^2$ denotes the coefficients for $j$-th term in the polynomial segment at $\omega_k$. For minimum snap trajectory optimization, $q=4$, objective function can be written as,
\textit{minimize} 
\begin{equation}
    \int_{0}^{t_{H}} {\norm{\sigma^{(q)}(t)}}^2 \textrm{d}t 
    \label{objective}
\end{equation}
\textit{subject to}
\begin{subequations}
  \begin{alignat*}{2}
    & \sigma (t_\textbf{k}) = \textbf{x}^{\textbf{k}*} \notag \\
    & \sigma ^{(q)} (t_\textbf{k}) = 0, \, \, \, 1\leq q \leq 4, \, \, \, (\textbf{k} = 0, \textbf{k} = H) \\
    & \sigma ^{(q)} (t_\textbf{k}) = 0, \, \, \, \textrm{or free,} \, \, \ 1\leq q \leq 4, \, 1 \leq \textbf{k} < H 
  \end{alignat*}
\end{subequations}

We constrain the optimization problem by introducing equality and inequality conditions on $\sigma^{(q)}(t)$, where $0 \leq q \leq 4$. Equality constraints for $\dot{\sigma}(t)$, $\ddot{\sigma}(t)$ and $\dddot{\sigma}(t)$ are necessary at each waypoint to ensure the continuity of velocity, acceleration and jerk profiles throughout the trajectory. The objective function Eq.~\eqref{objective} can be transformed into a quadratic program of the decision variable $\chi$ with a set of linear constraints as below.
\begin{equation} 
\label{qp}
    \begin{aligned}
    & \textit{minimize}
    & & \chi^TA\chi\\
    & \textit{subject to}
    & & B\chi \leq b
\end{aligned}
\end{equation}

The resulting decision vector contains the coefficients of the piecewise polynomials $\sigma(t)$, specifically for $[\sigma_X(t) \sigma_Y(t)]$ by dimensionalization. We use the optimal time allocation method introduced in \cite{liu2017planning} for refining the trajectories. The smooth trajectories are evaluated prior to the execution for collisions and convex corridor checks were added similar to \cite{mellinger2012trajectory} to limit the deviation of the trajectories from the original discrete paths. We use RHP to construct continuous smooth trajectories until the robots reach the final goal positions. Here, $t_H$ defines the {planning horizon}. With the continuous arrival of path nodes, the robot executes optimized trajectories up to the {execution horizon}. 
Thus, we are able to continuously navigate the robots as the MRF is being converged. Specially, this approach showed to be useful when optimizing over considerably large environments, as the distance to the goal position defines the total number of optimization steps of the MRF.

\section{Experiments and Results}

 \begin{figure*}[ht] \vspace{-10pt}
  \centering
  \subfigure[]
 	{\label{pent_nav} \includegraphics[width=0.23\textwidth,trim={1cm 1cm 1cm 1cm}, clip]{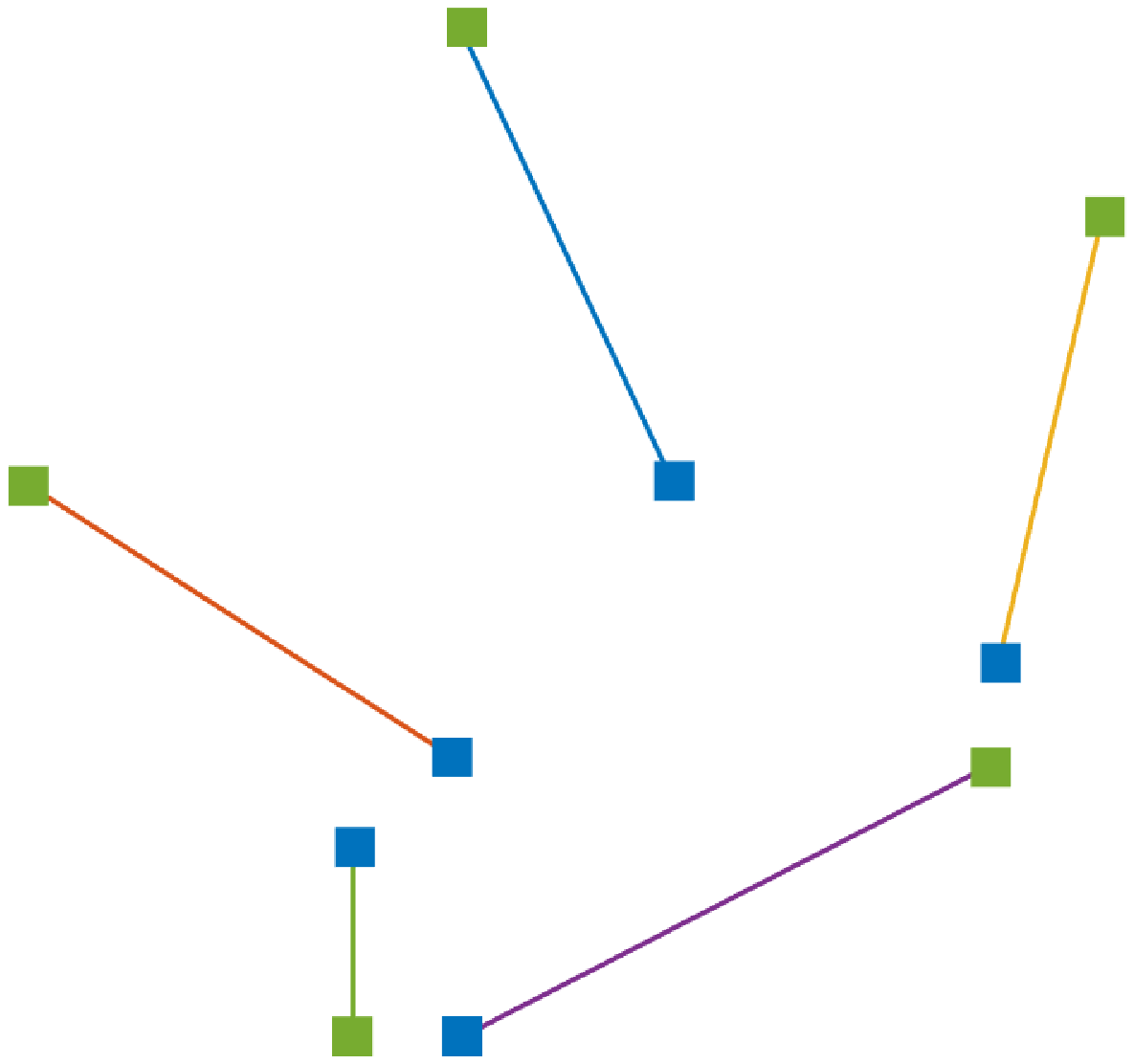}}
    \hfill
\subfigure[]
 	{\label{10_nav} \includegraphics[width=0.23\textwidth,trim={1cm 1cm 1cm 1cm}, clip]{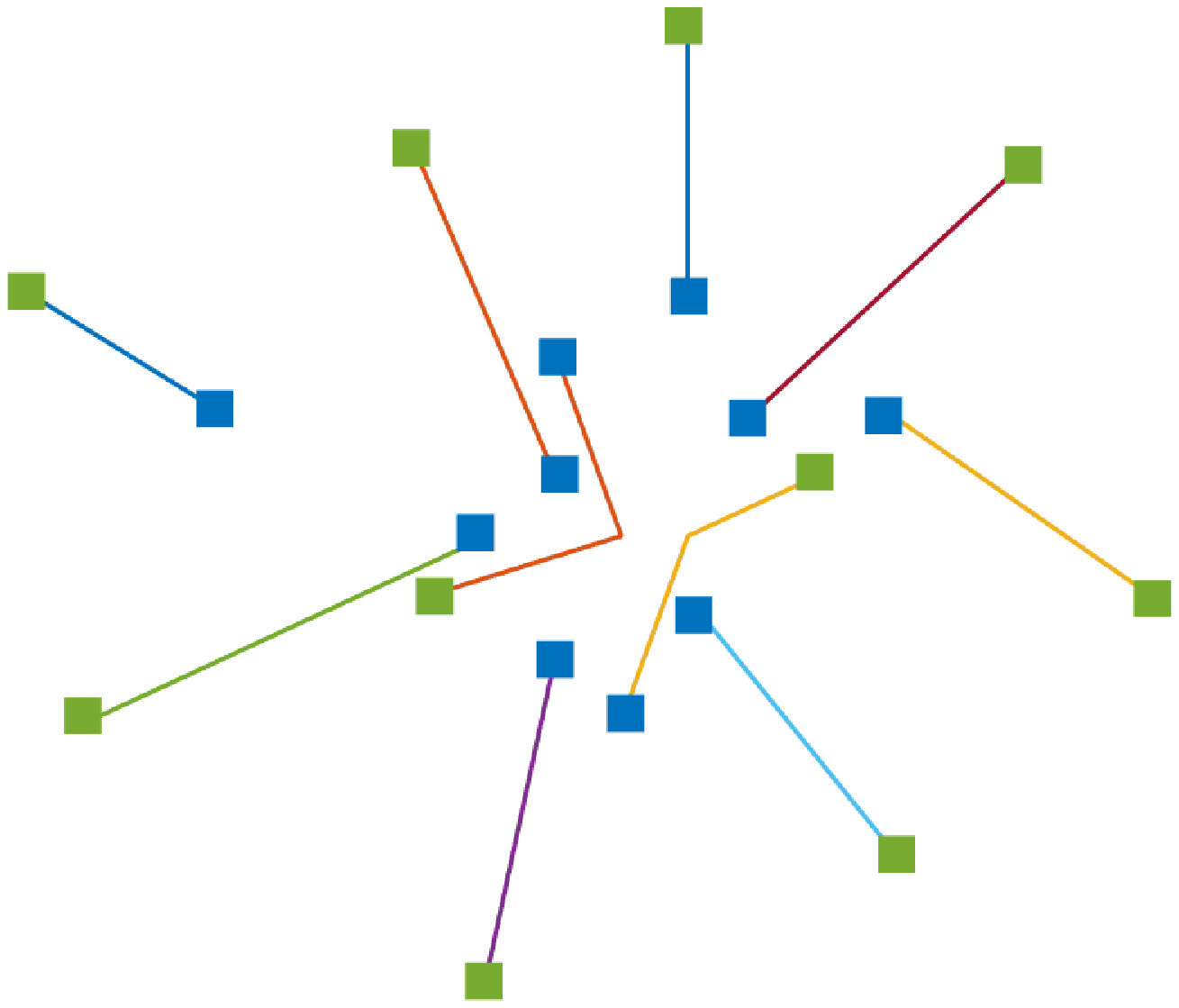}}
 	\hfill
\subfigure[]
 	{\label{corridor_nav} \includegraphics[width=0.22\textwidth,trim={1cm 1cm 1cm 1cm}, clip]{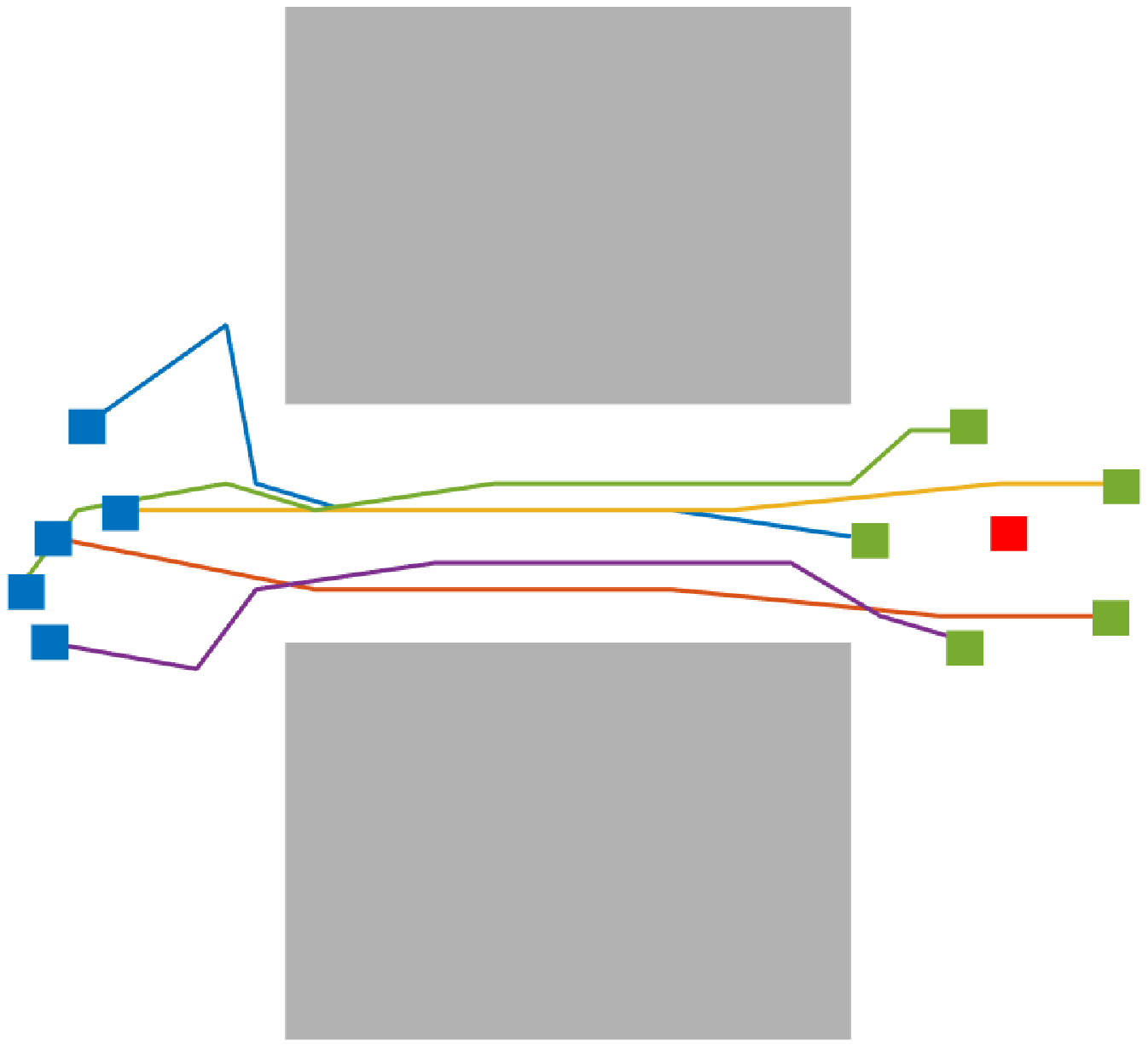}}
 \hfill
 \subfigure[]
 {\label{split_nav} \includegraphics[width=0.23\textwidth, trim={1cm 1cm 1cm 1cm}, clip]{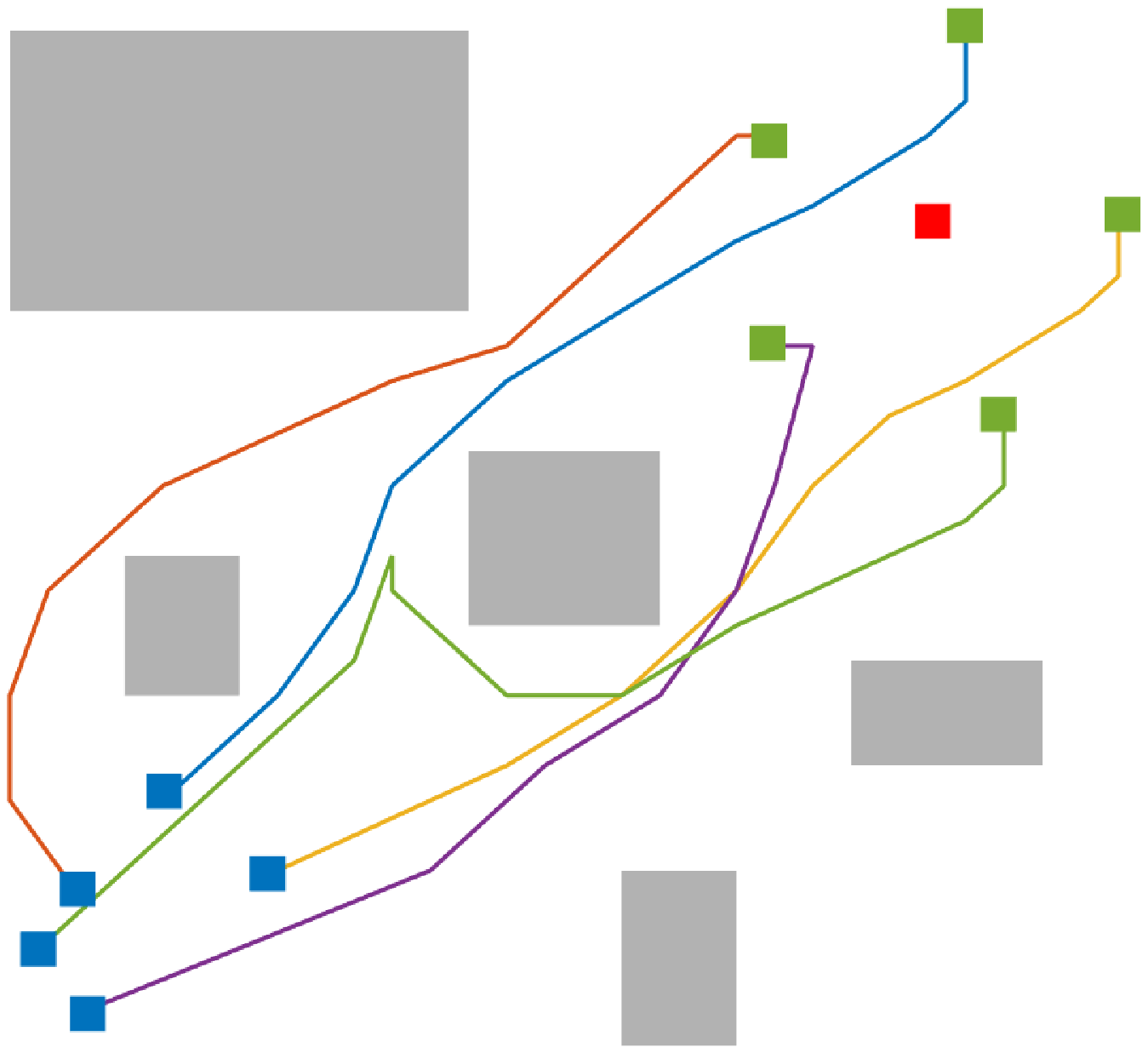}}
 
    \subfigure[]
 {\label{pent_opt} \includegraphics[width=0.23\textwidth,trim={1cm 1cm 1cm 1cm}, clip]{images/5_converged.eps}}
    \hfill
\subfigure[]
 	{\label{10_opt} \includegraphics[width=0.23\textwidth,trim={1cm 1cm 1cm 1cm}, clip]{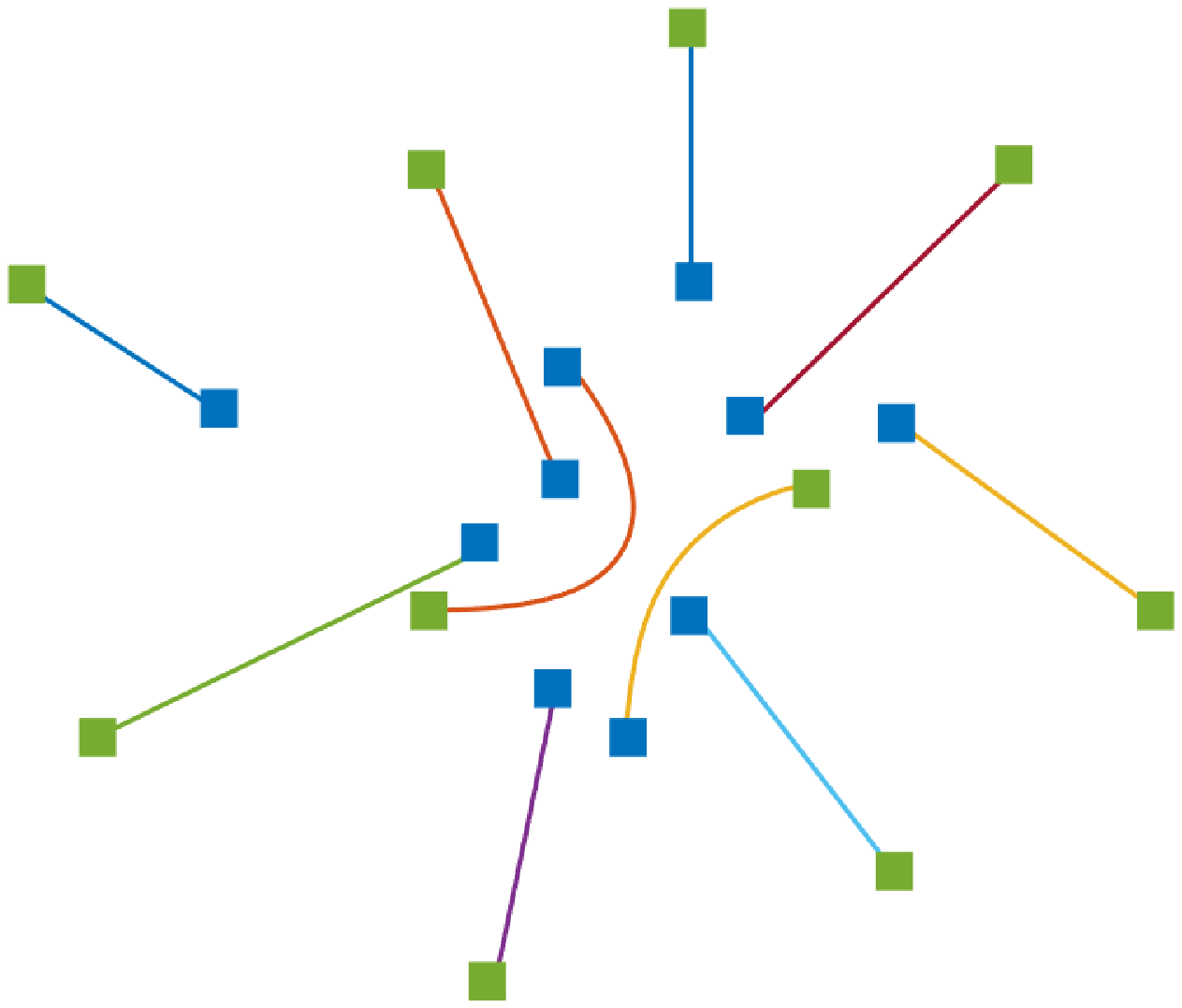}}
 	\hfill
\subfigure[]
 	{\label{corridor_opt} \includegraphics[width=0.22\textwidth,trim={1cm 1cm 1cm 1cm}, clip]{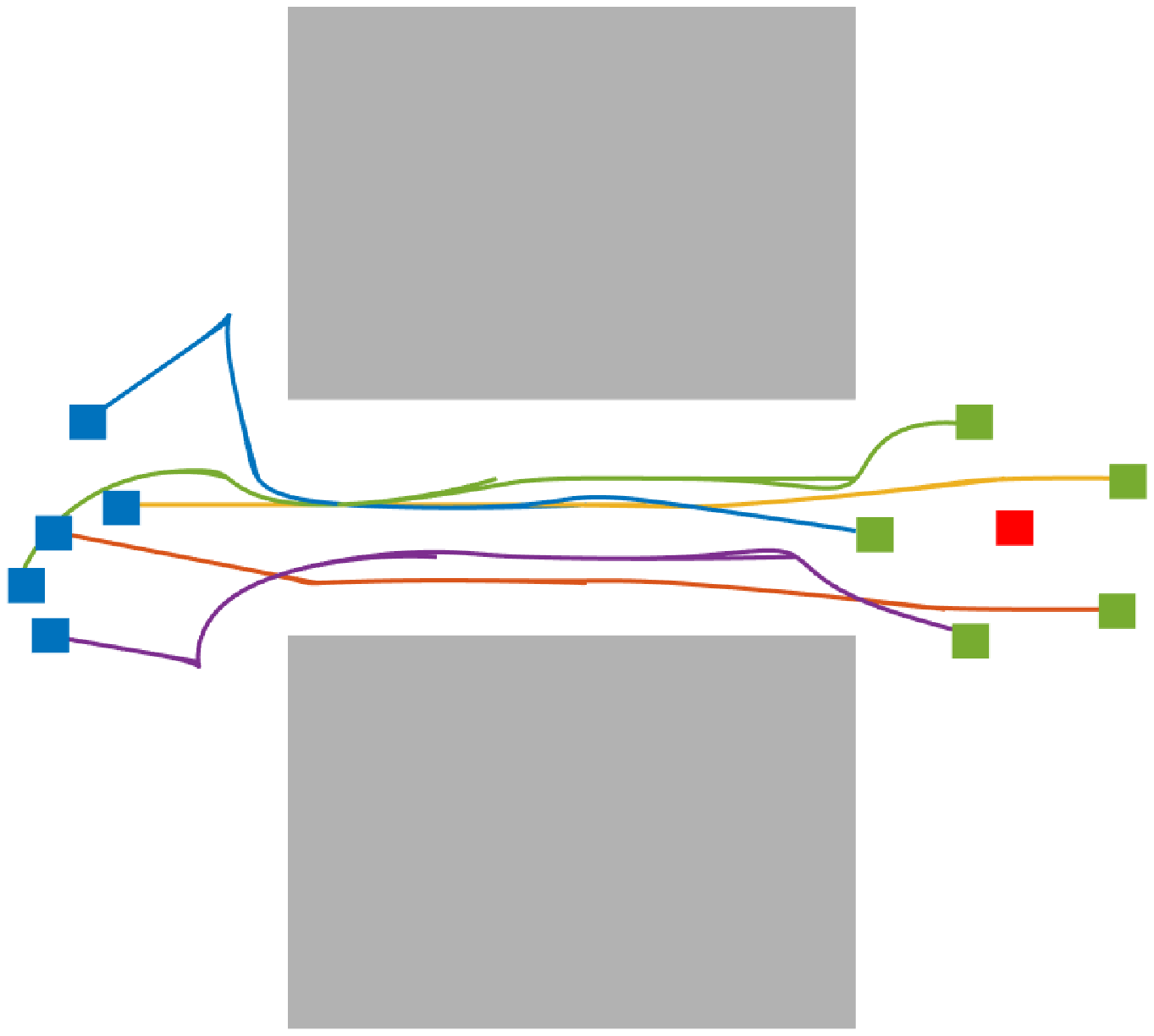}}
 \hfill\subfigure[]
 {\label{split_opt} \includegraphics[width=0.23\textwidth, trim={1cm 1cm 1cm 1.2cm}, clip]{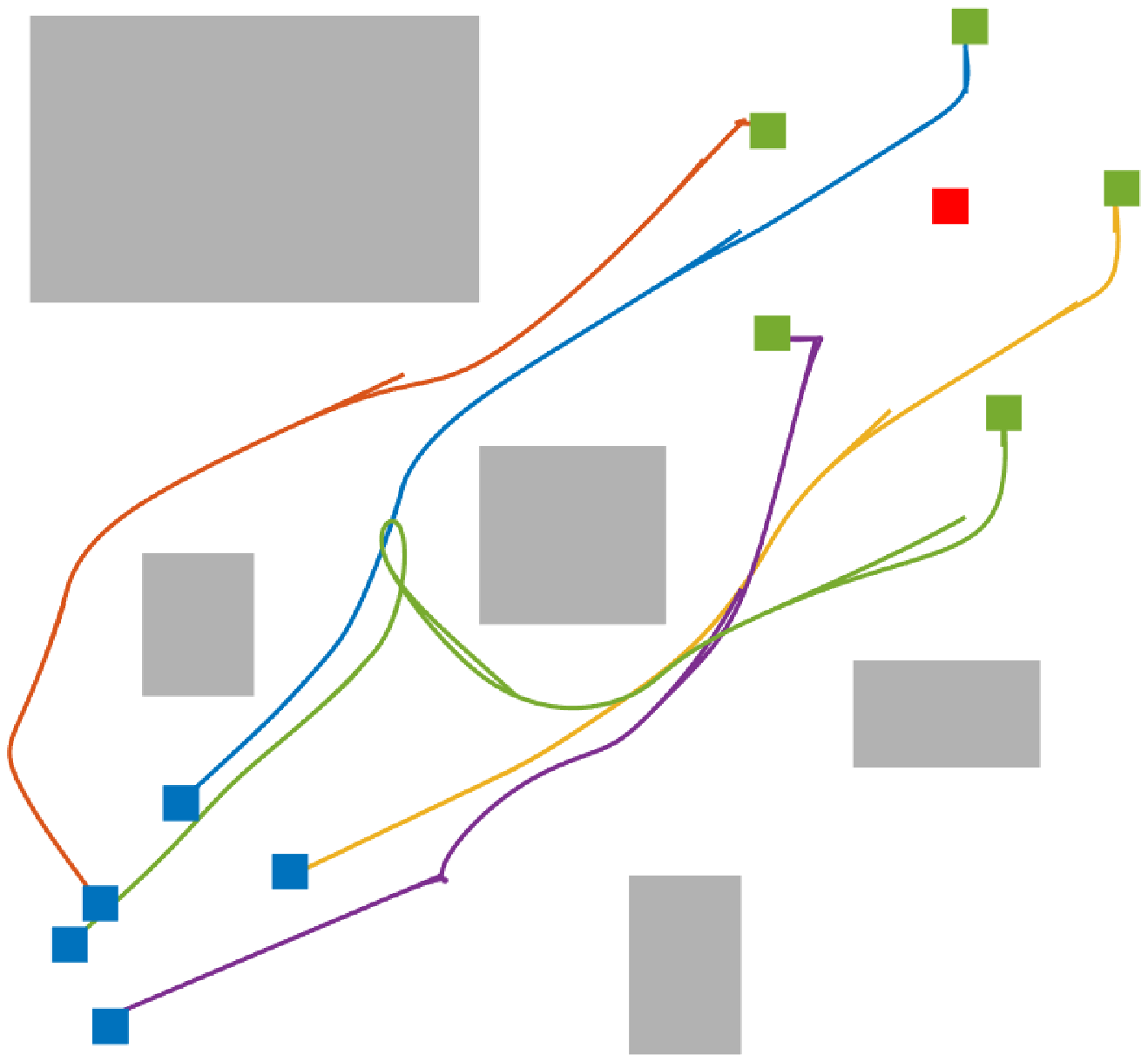}}
 
\caption{Final trajectories of swarms of 5 and 10 robots. Green and blue markers show the start and end positions of the robots respectively. Grey areas denote the obstacles. (a)(b) Groups of 5 and 10 robots converging to a minimum energy formation with $k=3$ and $k=7$. (c)(d) Groups of 5 robots navigating through a narrow corridor ($40m \times 40m$) and random blocks ($30m \times 30m$) ($k=3$). (e)(f) Optimized trajectories. (g)(h) Optimized trajectories with RHP. Sharp edges denote stop-start scenarios.} 
\vspace{-10pt}
\label{interaction_shapes}
\end{figure*}

We implemented the proposed approach on Matlab and simulated with quadrotors in a ROS/Gazebo environment. For multi-quadrotor navigation with trajectory optimization, we used the RHP framework proposed in \cite{fernando2019formation}. The maximal clique factorization of the interaction graph is performed using Bron-Kerbosch algorithm  \cite{bron1973algorithm} with pivoting. We discretized the environment into a 2D grid-map where each cell is $1m \times 1m$, assuming two adjacent cells can accommodate two robots without collisions. The initial robot positions are sampled from a multivariate Gaussian distribution with mean set to a fixed starting position for the robot team, by satisfying the necessary and sufficient condition $\forall i \in \mathcal{V}, \mathcal{N}_i \neq \varnothing.$ All the experiments were conducted on Matlab running on a Linux desktop with Intel core i7 CPU running at 3.6 GHz and 16Gb of memory. We used the following arbitrarily and experimentally chosen parameters throughout this section. The parameters on the interaction potential, $a_I$, $b_I$, $k_a$ and $k_r$ defines the cohesiveness of the swarm, yet changing them under the constraints introduced in \eqref{morse}, appeared to have a minimal effect in the convergence.

\begin{table}[ht]
\begin{tabular}{|l|l|l|}
\hline
\textbf{Potential Function} & \textbf{Parameter} & \textbf{Value} \\ \hline
Interaction potential       & $a_I$                & 0.7               \\ \cline{2-3} 
                   & $b_I$                & 0.9               \\ \cline{2-3} 
                   & $k_a$              & 14               \\ \cline{2-3} 
                   & $k_r$              & 4               \\ \hline
Goal potential               & $a_G$                  & 3              \\ \cline{2-3} 
                   & $k_g$              & 20             \\ \hline
Obstacle potential           & $\hat{\sigma}$     & 1      \\ \cline{2-3}
                   & $\Gamma$          & 5 \\ \hline
\end{tabular}
\end{table}


\begin{figure}[ht]
\subfigure[]
 	{\label{pos_plot} \includegraphics[scale=0.3,trim={0.1cm 0cm 1.2cm 0cm}, clip]{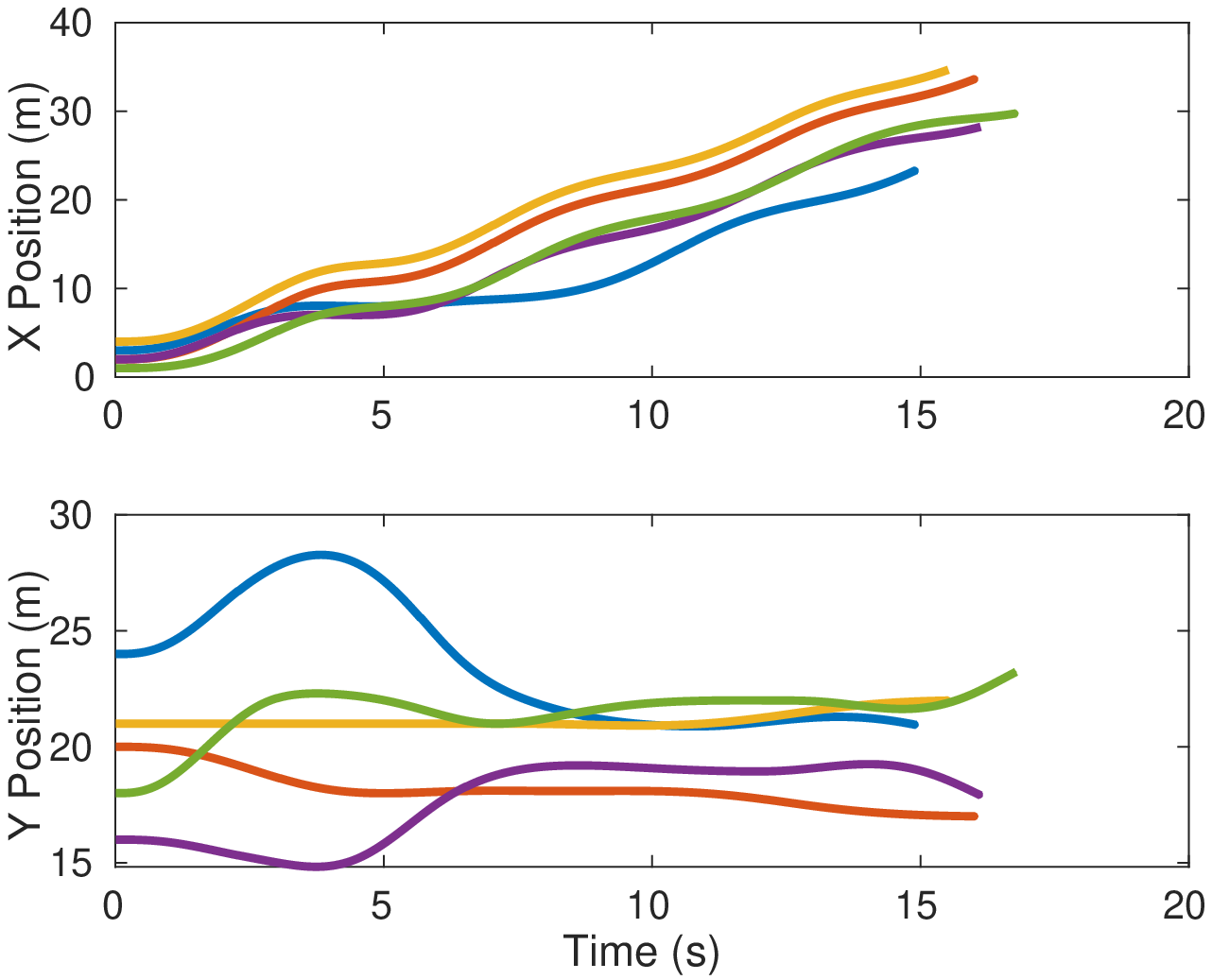}}
 \subfigure[]
 	{\label{swarm_size} \includegraphics[scale=0.3,trim={0cm 0cm 1.2cm 0cm}, clip]{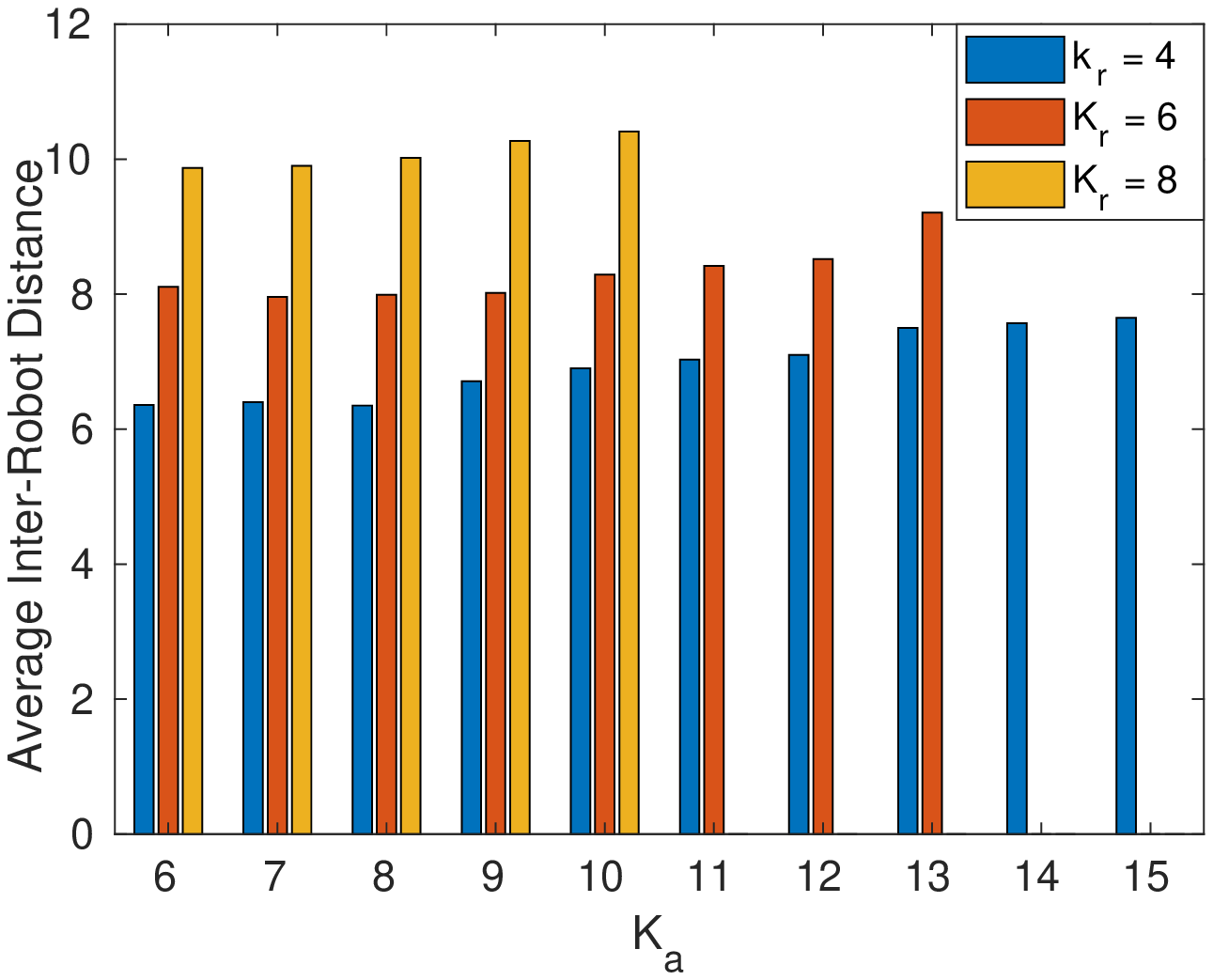}}
 \subfigure[]
 	{\label{distance_plot} \includegraphics[width=0.98\textwidth,trim={0.5cm 0cm 0cm 0cm}, clip]{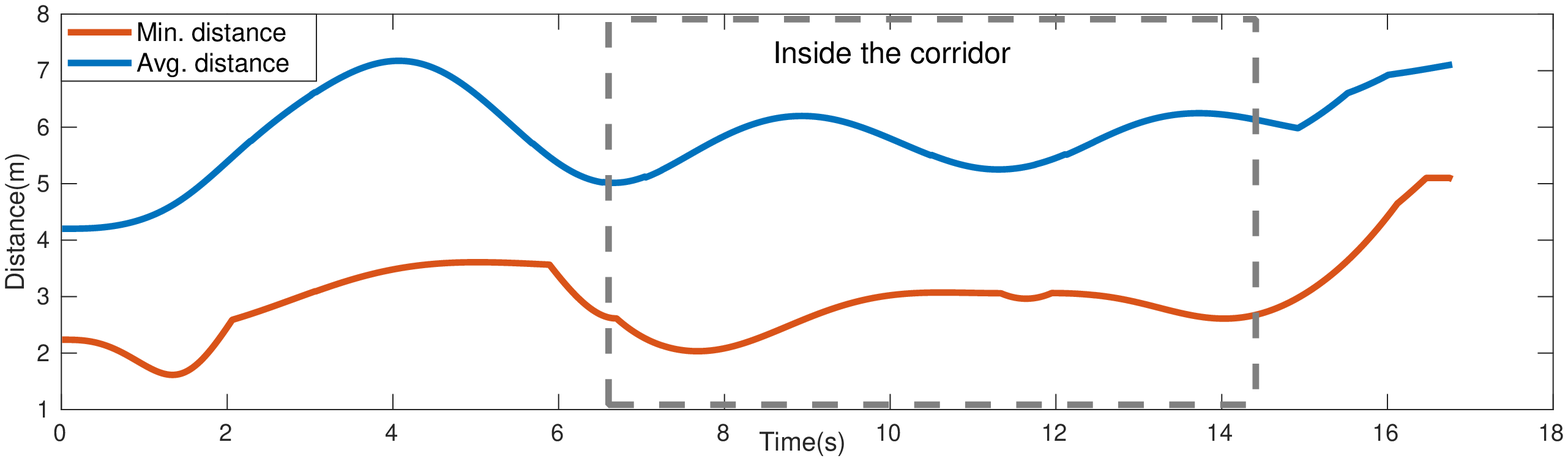}}
\caption{(a) Trajectories of individual robots. (b) Swarm size against $k_a$ and $k_r$. (c) Minimum and average distances between the robots for narrow corridor. }
\end{figure}

\begin{figure}[ht]
 \subfigure[]
 	{\label{convergence} \includegraphics[scale=0.3,trim={0.5cm 0cm 1.2cm 0cm}, clip]{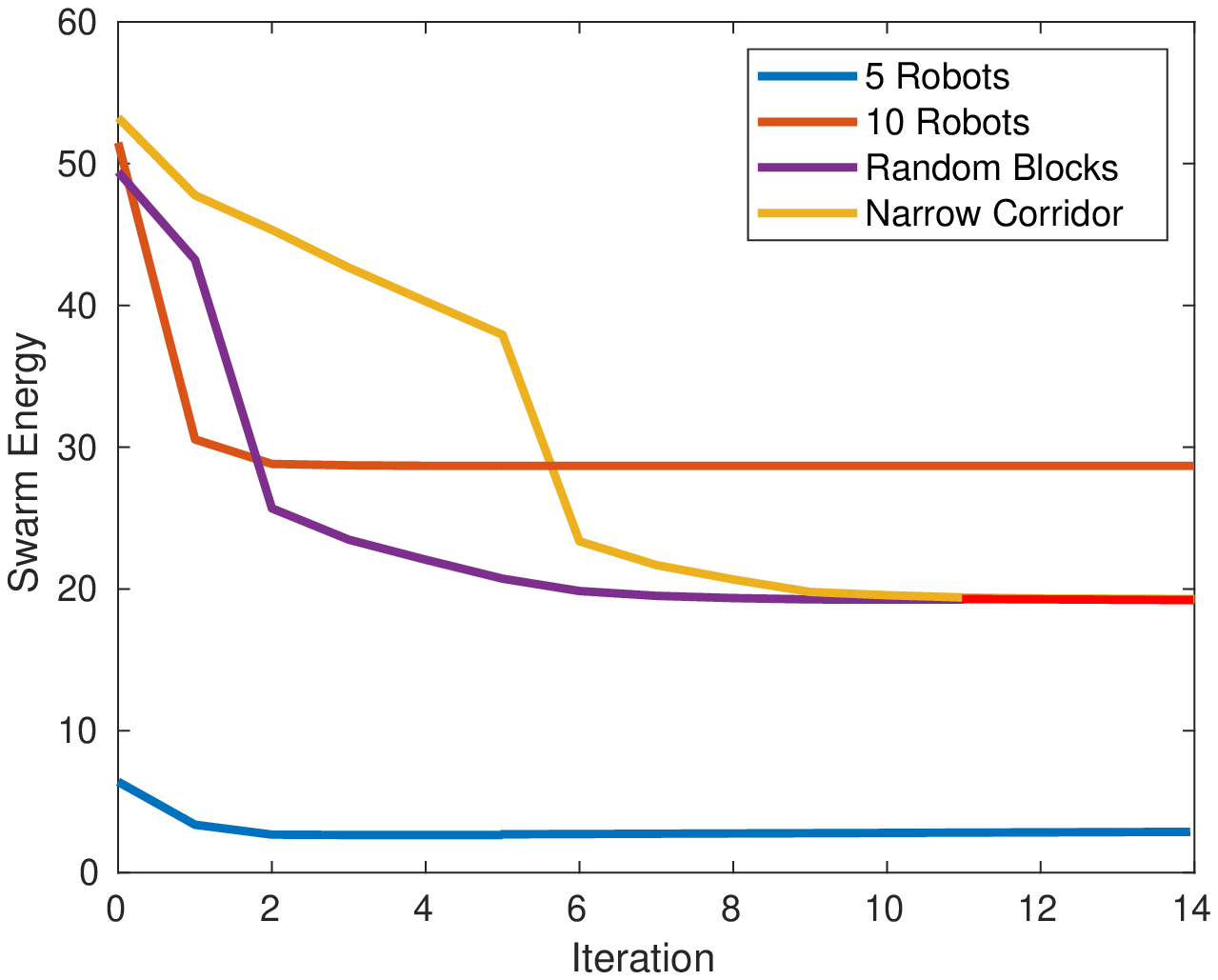}}
\subfigure[]
 	{\label{times} \includegraphics[scale=0.3,trim={0.5cm 0cm 1.2cm 0cm}, clip]{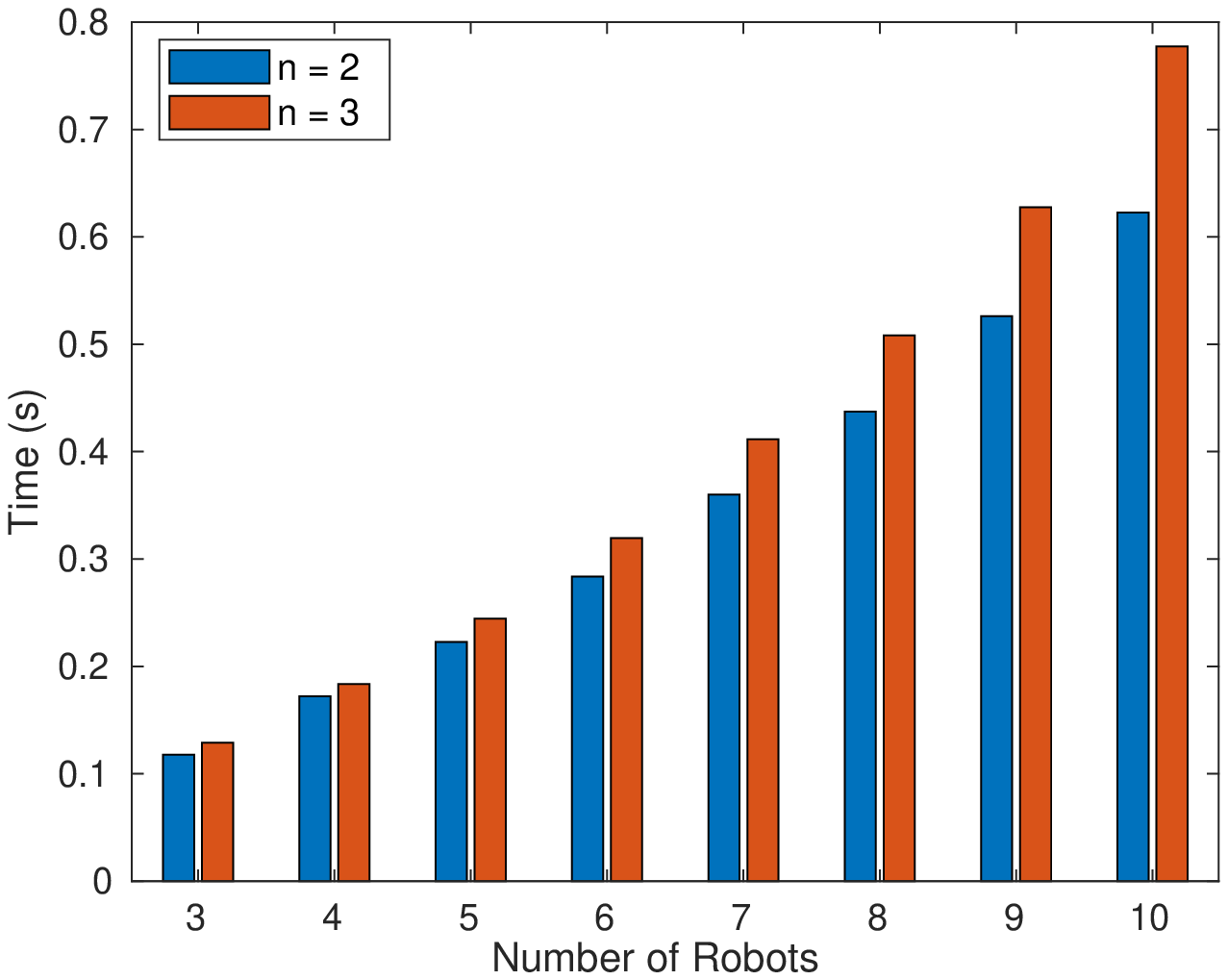}}
\caption{(a) Convergence plot for the 4 scenarios. (b) Time per MRF optimization iteration in Matlab. \vspace{-10pt}} 
\end{figure}

First we experimented the convergence of multiple robots in absence of external potential fields, by setting $\rho_S = 0$ in Eq. \eqref{energy}. Fig. \ref{pent_nav}, \ref{10_nav} show the initial and converged configurations with discrete trajectories for 5 and 10 robot configurations. It can be observed that the robots tend to maintain a similar distance to each other in order to balance the repulsion and the attraction forces in such scenarios. By integrating external potential fields, we tested our approach for narrow corridor and random blocks scenarios as in Fig. \ref{corridor_nav} and \ref{split_nav}. We observed that under relatively small $k_a$ and large $a_i$ values, robots are capable of maintaining a tight cohesive formations such as pentagons, at each MRF optimization horizons. However, this can result in collisions in between the horizons, in tighter environments unless explicitly handled in the trajectory optimization phase. Further, due to the topological neighborhood, we observed that the cohesive behavior in large swarms lead the robots to wait until the other robots get closer or mover over, similar to natural aggregations. 

Convergence of the swarm energy throughout the MRF optimization process is showed in Fig \ref{convergence}. Clearly, the optimization for free space scenarios completed under 3 optimization iterations resulting discrete trajectories with minimal edges. However, in corridor and random blocks scenarios, the number of optimization steps depend on the distance to the goal position and the order of the local search space. Even though, this poses a minimal effect on long distance robot team navigation with RHP, the number of connection points can affect the performances of the trajectory optimization. Therefore, we applied a waypoint pruning method as discussed earlier to reduce the number of rough edges in discrete paths. In our experiments we used 4 local optimization steps as the planning horizon for both waypoint pruning and RHP. The optimized trajectories for each scenario is depicted in Fig. \ref{pent_opt}, \ref{10_opt}, \ref{corridor_opt} and \ref{split_opt}. During the trajectory optimization, each optimized trajectory is computed with stationary start and end states. This approach is useful when; (a) handling stop-start scenarios, and (b) handling the emergencies during the navigation. We used 50\% as the {execution horizon} in RHP. Sharp edges in the optimized trajectories denote the stop-start scenarios where robots are waiting till other robots to move over. 
We use the average distance between the robots as a consensus to denote the cohesiveness of the swarm. Fig. \ref{distance_plot} shows the variability of the minimum and the average distances between the robots during the navigation through a narrow corridor. The first peak from 2s to 4s, can be accounted to the initial \textit{expansion} where the robots try to move away from each other due to a tight initialization. Individual robot trajectories during the travel is shown in Fig. \ref{pos_plot}. We observed that the swarm size varies with $k_r$ and $k_a$ when $\frac{b_Ik_r}{a_Ik_a} < 1$. Fig. \ref{swarm_size} shows the change of the average inter-robot distance for different $k_a$ and $k_r$ values while keeping $a_I$, $b_I$ constant. We observed that smaller $k_a$ and $a_I$ values can generate tighter swarm formations \ref{swarm_size}, which can be useful to maintain swarm shapes in narrow environments, whereas larger values are much robust in splitting and merging configurations. However, we noted that a large range of values can perform in complex environments that encompass both the scenarios. The computational time for optimizing a single horizon using ICM is shown in \ref{times} against the number of robots and the order of the neighborhood. 

\section{Conclusion 
}
In this work we propose a novel approach to navigate a team of robots in obstacle environments via dynamic MRF optimization. We model the swarm formation dynamics on artificial potentials that exert on the robots. By locally minimizing the swarm energy, a set of collision-free paths are computed and further smoothed using trajectory optimization. We experimented the approach on a group of aerial robots on Gazebo simulation environment with a receding horizon planning paradigm. We show that this approach can be used to generate dynamically feasible trajectories for robots in complex environments while mimicking natural like cohesive behaviors. In addition, our approach does not require specifying individual goal positions or interim transitional formations. 


\bibliographystyle{unsrt}
\bibliography{root}

\begin{thebibliography}{10}

\bibitem{cardona2019robot}
Gustavo~A Cardona and Juan~M Calderon.
\newblock Robot swarm navigation and victim detection using rendezvous
  consensus in search and rescue operations.
\newblock {\em Applied Sciences}, 9(8):1702, 2019.

\bibitem{honig2018trajectory}
Wolfgang H{\"o}nig, James~A Preiss, TK~Satish Kumar, Gaurav~S Sukhatme, and
  Nora Ayanian.
\newblock Trajectory planning for quadrotor swarms.
\newblock {\em IEEE Transactions on Robotics}, (99):1--14, 2018.

\bibitem{cappo2018online}
Ellen~A Cappo, Arjav Desai, Matthew Collins, and Nathan Michael.
\newblock Online planning for human--multi-robot interactive theatrical
  performance.
\newblock {\em Autonomous Robots}, 42(8):1771--1786, 2018.

\bibitem{turpin2013trajectory}
Matthew Turpin, Nathan Michael, and Vijay Kumar.
\newblock Trajectory planning and assignment in multirobot systems.
\newblock In {\em Algorithmic foundations of robotics X}, pages 175--190.
  Springer, 2013.

\bibitem{michael2011control}
Nathan Michael and Vijay Kumar.
\newblock Control of ensembles of aerial robots.
\newblock {\em Proceedings of the IEEE}, 99(9):1587--1602, 2011.

\bibitem{desai2001modeling}
Jaydev~P Desai, James~P Ostrowski, and Vijay Kumar.
\newblock Modeling and control of formations of nonholonomic mobile robots.
\newblock {\em IEEE transactions on Robotics and Automation}, 17(6):905--908,
  2001.

\bibitem{shishika2017mosquito}
Daigo Shishika and Derek~A Paley.
\newblock Mosquito-inspired swarming for decentralized pursuit with autonomous
  vehicles.
\newblock In {\em 2017 American Control Conference (ACC)}, pages 923--929.
  IEEE, 2017.

\bibitem{ze2012formation}
Cai Ze-Su, Zhao Jie, and Cao Jian.
\newblock Formation control and obstacle avoidance for multiple robots subject
  to wheel-slip.
\newblock {\em International Journal of Advanced Robotic Systems}, 9(5):188,
  2012.

\bibitem{gayle2009multi}
Russell Gayle, William Moss, Ming~C Lin, and Dinesh Manocha.
\newblock Multi-robot coordination using generalized social potential fields.
\newblock In {\em 2009 IEEE International Conference on Robotics and
  Automation}, pages 106--113. IEEE, 2009.

\bibitem{okubo1986dynamical}
Akira Okubo.
\newblock Dynamical aspects of animal grouping: swarms, schools, flocks, and
  herds.
\newblock {\em Advances in biophysics}, 22:1--94, 1986.

\bibitem{sanchez2002delaying}
Gildardo S{\'a}nchez and Jean-Claude Latombe.
\newblock On delaying collision checking in prm planning: Application to
  multi-robot coordination.
\newblock {\em The International Journal of Robotics Research}, 21(1):5--26,
  2002.

\bibitem{guo2002distributed}
Yi~Guo and Lynne~E Parker.
\newblock A distributed and optimal motion planning approach for multiple
  mobile robots.
\newblock In {\em Proceedings 2002 IEEE International Conference on Robotics
  and Automation (Cat. No. 02CH37292)}, volume~3, pages 2612--2619. IEEE, 2002.

\bibitem{yu2016optimal}
Jingjin Yu and Steven~M LaValle.
\newblock Optimal multirobot path planning on graphs: Complete algorithms and
  effective heuristics.
\newblock {\em IEEE Transactions on Robotics}, 32(5):1163--1177, 2016.

\bibitem{chuang2007multi}
Yao-Li Chuang, Yuan~R Huang, Maria~R D'Orsogna, and Andrea~L Bertozzi.
\newblock Multi-vehicle flocking: scalability of cooperative control algorithms
  using pairwise potentials.
\newblock In {\em Proceedings 2007 IEEE international conference on robotics
  and automation}, pages 2292--2299. IEEE, 2007.

\bibitem{nazarahari2019multi}
Milad Nazarahari, Esmaeel Khanmirza, and Samira Doostie.
\newblock Multi-objective multi-robot path planning in continuous environment
  using an enhanced genetic algorithm.
\newblock {\em Expert Systems with Applications}, 115:106--120, 2019.

\bibitem{bennet2010distributed}
Derek~J Bennet and Colin~R McInnes.
\newblock Distributed control of multi-robot systems using bifurcating
  potential fields.
\newblock {\em Robotics and Autonomous Systems}, 58(3):256--264, 2010.

\bibitem{toksoz2019decentralized}
Mehmet~Altan Toks{\"o}z, Sinan O{\u{g}}uz, and Veysel Gazi.
\newblock Decentralized formation control of a swarm of quadrotor helicopters.
\newblock In {\em 2019 IEEE 15th International Conference on Control and
  Automation (ICCA)}, pages 1006--1013. IEEE, 2019.

\bibitem{augugliaro2012generation}
Federico Augugliaro, Angela~P Schoellig, and Raffaello D'Andrea.
\newblock Generation of collision-free trajectories for a quadrocopter fleet: A
  sequential convex programming approach.
\newblock In {\em Intelligent Robots and Systems (IROS), 2012 IEEE/RSJ
  International Conference on}, pages 1917--1922. IEEE, 2012.

\bibitem{alonso2017multi}
Javier Alonso-Mora, Stuart Baker, and Daniela Rus.
\newblock Multi-robot formation control and object transport in dynamic
  environments via constrained optimization.
\newblock {\em The International Journal of Robotics Research},
  36(9):1000--1021, 2017.

\bibitem{baras2004control}
John~S Baras and Xiaobo Tan.
\newblock Control of autonomous swarms using gibbs sampling.
\newblock In {\em 2004 43rd IEEE Conference on Decision and Control (CDC)(IEEE
  Cat. No. 04CH37601)}, volume~5, pages 4752--4757. IEEE, 2004.

\bibitem{xi2006gibbs}
Wei Xi, Xiaobo Tan, and John~S Baras.
\newblock Gibbs sampler-based coordination of autonomous swarms.
\newblock {\em Automatica}, 42(7):1107--1119, 2006.

\bibitem{spears2004distributed}
William~M Spears, Diana~F Spears, Jerry~C Hamann, and Rodney Heil.
\newblock Distributed, physics-based control of swarms of vehicles.
\newblock {\em Autonomous Robots}, 17(2-3):137--162, 2004.

\bibitem{bishop2006pattern}
Christopher~M Bishop.
\newblock {\em Pattern recognition and machine learning}.
\newblock springer, 2006.

\bibitem{ballerini2008interaction}
Michele Ballerini, Nicola Cabibbo, Raphael Candelier, Andrea Cavagna, Evaristo
  Cisbani, Irene Giardina, Vivien Lecomte, Alberto Orlandi, Giorgio Parisi,
  Andrea Procaccini, et~al.
\newblock Interaction ruling animal collective behavior depends on topological
  rather than metric distance: Evidence from a field study.
\newblock {\em Proceedings of the national academy of sciences},
  105(4):1232--1237, 2008.

\bibitem{gazi2013lagrangian}
Veysel Gazi.
\newblock On lagrangian dynamics based modeling of swarm behavior.
\newblock {\em Physica D: Nonlinear Phenomena}, 260:159--175, 2013.

\bibitem{d2006self}
Maria~R D’Orsogna, Yao-Li Chuang, Andrea~L Bertozzi, and Lincoln~S Chayes.
\newblock Self-propelled particles with soft-core interactions: patterns,
  stability, and collapse.
\newblock {\em Physical review letters}, 96(10):104302, 2006.

\bibitem{khatib1986real}
Oussama Khatib.
\newblock Real-time obstacle avoidance for manipulators and mobile robots.
\newblock In {\em Autonomous robot vehicles}, pages 396--404. Springer, 1986.

\bibitem{fix2011graph}
Alexander Fix, Aritanan Gruber, Endre Boros, and Ramin Zabih.
\newblock A graph cut algorithm for higher-order markov random fields.
\newblock In {\em 2011 International Conference on Computer Vision}, pages
  1020--1027. IEEE, 2011.

\bibitem{kohli2008partial}
Pushmeet Kohli, Alexander Shekhovtsov, Carsten Rother, Vladimir Kolmogorov, and
  Philip Torr.
\newblock On partial optimality in multi-label mrfs.
\newblock In {\em Proceedings of the 25th international conference on Machine
  learning}, pages 480--487, 2008.

\bibitem{ishikawa2010transformation}
Hiroshi Ishikawa.
\newblock Transformation of general binary mrf minimization to the first-order
  case.
\newblock {\em IEEE transactions on pattern analysis and machine intelligence},
  33(6):1234--1249, 2010.

\bibitem{richter2016polynomial}
Charles Richter, Adam Bry, and Nicholas Roy.
\newblock Polynomial trajectory planning for aggressive quadrotor flight in
  dense indoor environments.
\newblock In {\em Robotics Research}, pages 649--666. Springer, 2016.

\bibitem{liu2017planning}
Sikang Liu, Michael Watterson, Kartik Mohta, Ke~Sun, Subhrajit Bhattacharya,
  Camillo~J Taylor, and Vijay Kumar.
\newblock Planning dynamically feasible trajectories for quadrotors using safe
  flight corridors in 3-d complex environments.
\newblock {\em IEEE Robotics and Automation Letters}, 2(3):1688--1695, 2017.

\bibitem{mellinger2012trajectory}
Daniel Mellinger, Nathan Michael, and Vijay Kumar.
\newblock Trajectory generation and control for precise aggressive maneuvers
  with quadrotors.
\newblock {\em The International Journal of Robotics Research}, 31(5):664--674,
  2012.

\bibitem{fernando2019formation}
Malintha Fernando and Lantao Liu.
\newblock Formation control and navigation of a quadrotor swarm.
\newblock In {\em 2019 International Conference on Unmanned Aircraft Systems
  (ICUAS)}, pages 284--291. IEEE, 2019.

\bibitem{bron1973algorithm}
Coen Bron and Joep Kerbosch.
\newblock Algorithm 457: finding all cliques of an undirected graph.
\newblock {\em Communications of the ACM}, 16(9):575--577, 1973.

\end{thebibliography}
\end{document}